\documentclass[manuscript,screen,natbib=false]{acmart}

\usepackage{amsfonts}
\usepackage{graphicx, float}
\usepackage{wrapfig}
\usepackage{soul}
\usepackage{acro}

\usepackage{textcomp}
\usepackage{hyperref}

\usepackage{indentfirst}

\usepackage[backend=biber,sorting=none,doi=false,isbn=false,url=false,eprint=false,style=trad-abbrv]{biblatex}
\newcommand{\makeDatasetComparisonTable}{%
\begin{table}[!htbp]
    \centering
    \includegraphics[width=0.5\linewidth]{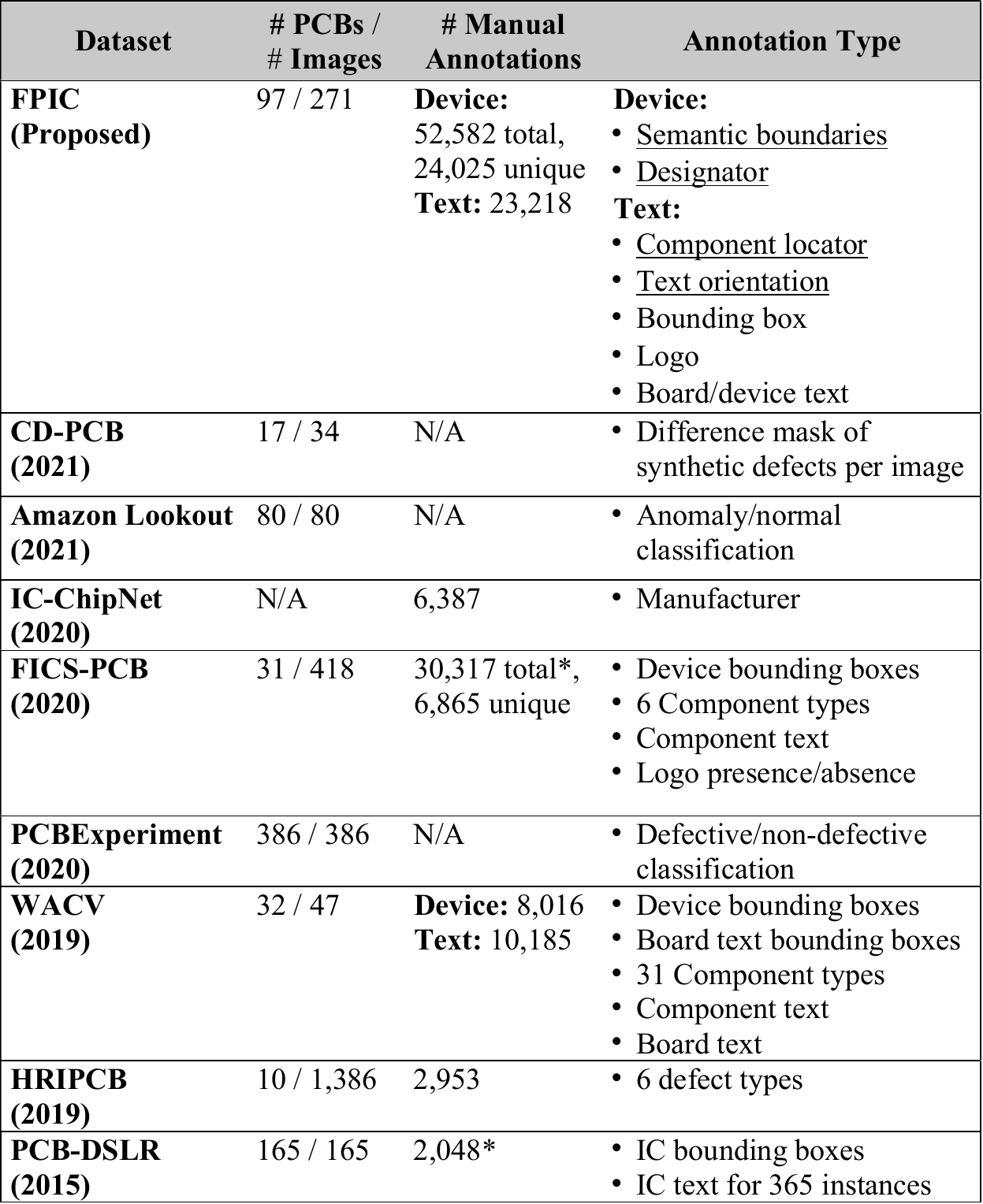}
    \caption[Summary of \ac{pcb} datasets in \ac{aoi} / \ac{ml} literature.]{Summary of \ac{pcb} datasets in \ac{aoi} / machine learning literature.
    Underlined fields are annotation types introduced for the first time in \ac{pcb} dataset literature.
    
    * More annotations are present in the downloadable dataset, but they are simple geometric transforms of the number stated in this table. FICS-PCB lists 77,347 annotations but has several copies of the same microscope annotations applied to different imaging conditions. PCB-DSLR lists 9,313 annotations but uses several rotated versions of the same 2,048 annotations.
    }
    \label{tab:dataset_comparison}
\end{table}
}

\newcommand{\makeDatasetFoldersTable}{%
\begin{table}[!htbp]
	\centering
	\includegraphics[width=0.8\linewidth]{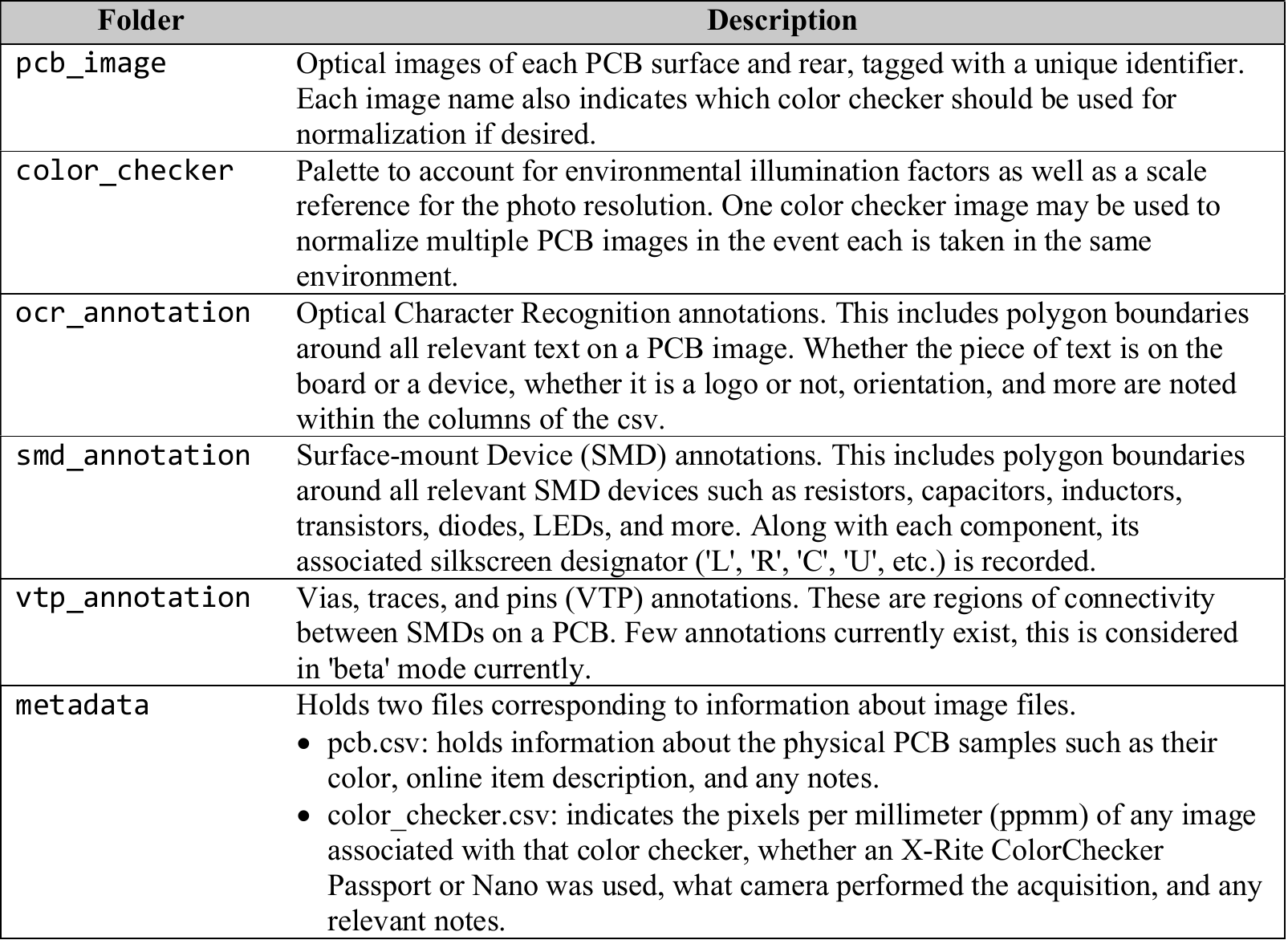}
	\caption{Description of each folder present in the Trust-Hub hosted \ac{fpic} dataset. All annotation and metadata are stored as CSV files, while images are in PNG format.}
	\label{tab:dataset_folders}
\end{table}
}

\newcommand{\makeAoiPictoralOverviewFig}{%
\begin{figure}[!htbp]
	\centering
	\includegraphics[width=\linewidth]{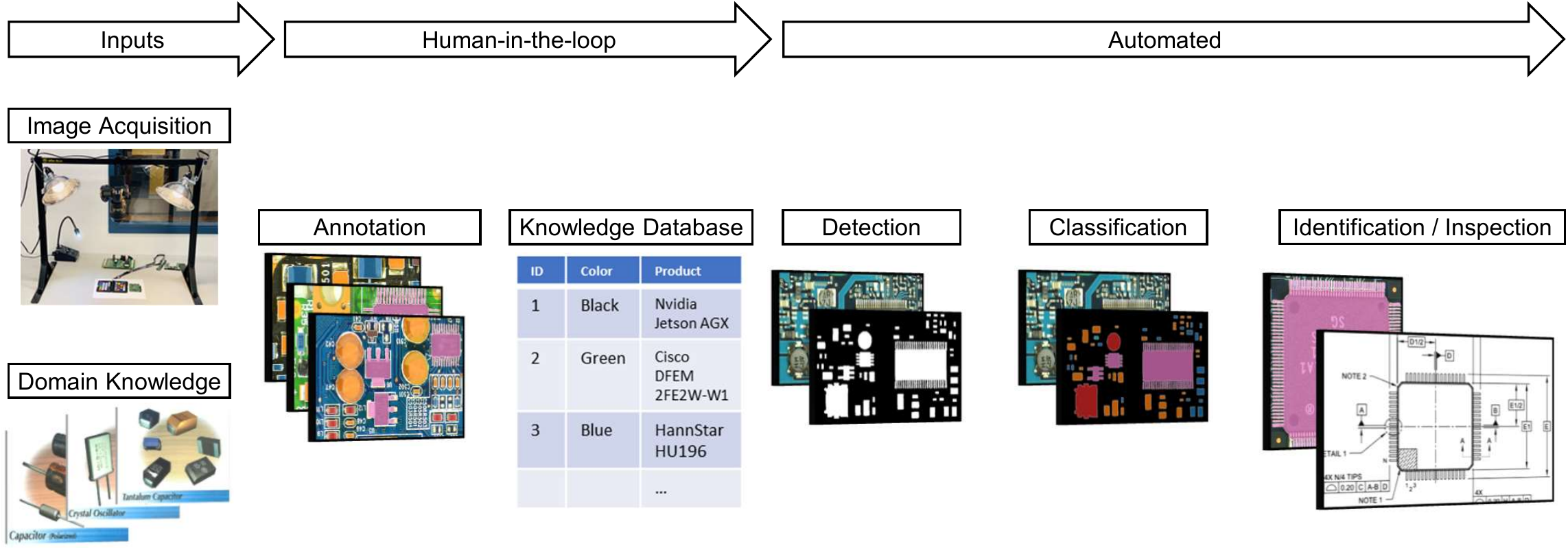}
	\caption{Overview of the stages present in most \ac{pcba} assurance workflows. Note the strong dependence on a wide variety of database samples and domain knowledge / experience.}
	\label{fig:aoi_pictoral_overview}
\end{figure}
}

\newcommand{\makeBboxVsSemanticFig}{%
\begin{figure}[!htbp]
	\centering
	\includegraphics[width=0.5\linewidth]{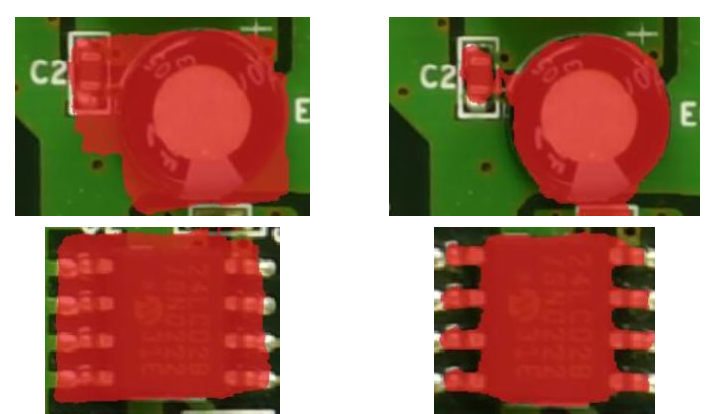}
	\caption{Comparison of predictions of a bounding-box trained LinkNet architecture vs. semantic masks. In the former case, regions next to each other were considered the same component while they were correctly distinguished in the latter.}
	\label{fig:bbox_vs_semantic}
\end{figure}
}

\newcommand{\makeBomPropertyExtractionFig}{%
\begin{figure}[!htbp]
	\centering
	\includegraphics[width=\linewidth]{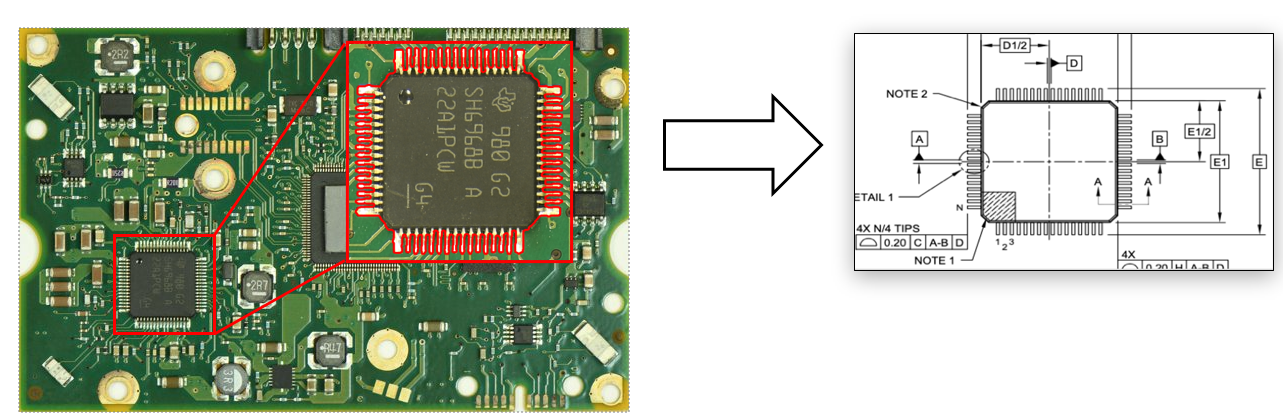}
	\caption{Accurate semantic boundaries drastically ease the process of extracting relevant BoM characteristics of various components.}
	\label{fig:bom_property_extraction}
\end{figure}	
}

\newcommand{\makeSimilarLookingSmdsFig}{%
\begin{figure}[!htbp]
	\centering
	\includegraphics[width=\linewidth]{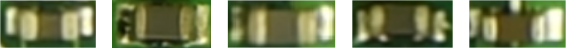}
	\caption{A thermistor, two inductors, ferrite bead, and a capacitor which all share similar visual characteristics. Non-\ac{sme} annotators may easily confuse the component type without assistance from silkscreen designators.}
	\label{fig:similar_looking_smds}
\end{figure}
}

\newcommand{\makeAnnotatedImageFig}{%
\begin{figure}[!htbp]
	\centering
	\includegraphics[width=0.75\linewidth]{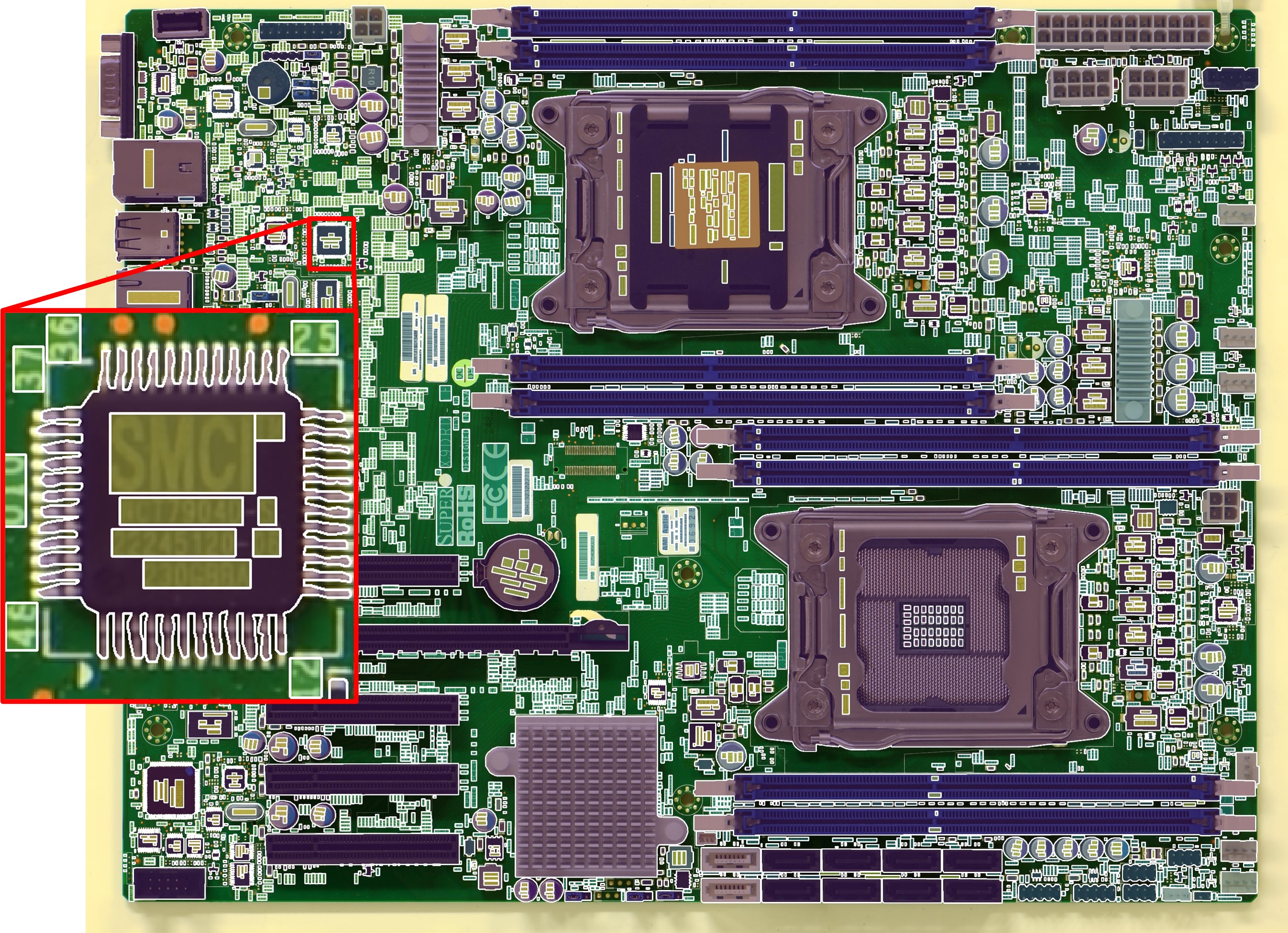}
	\caption{Sample annotated \ac{pcb}. The image has roughly 4,000 text instances and 1,500 \ac{smd} instances.}
	\label{fig:annotated_image}
\end{figure}
}

\newcommand{\makeSampleAnnotatedComponentFig}{%
\begin{figure}[!htbp]
	\centering
	\includegraphics[width=0.5\linewidth]{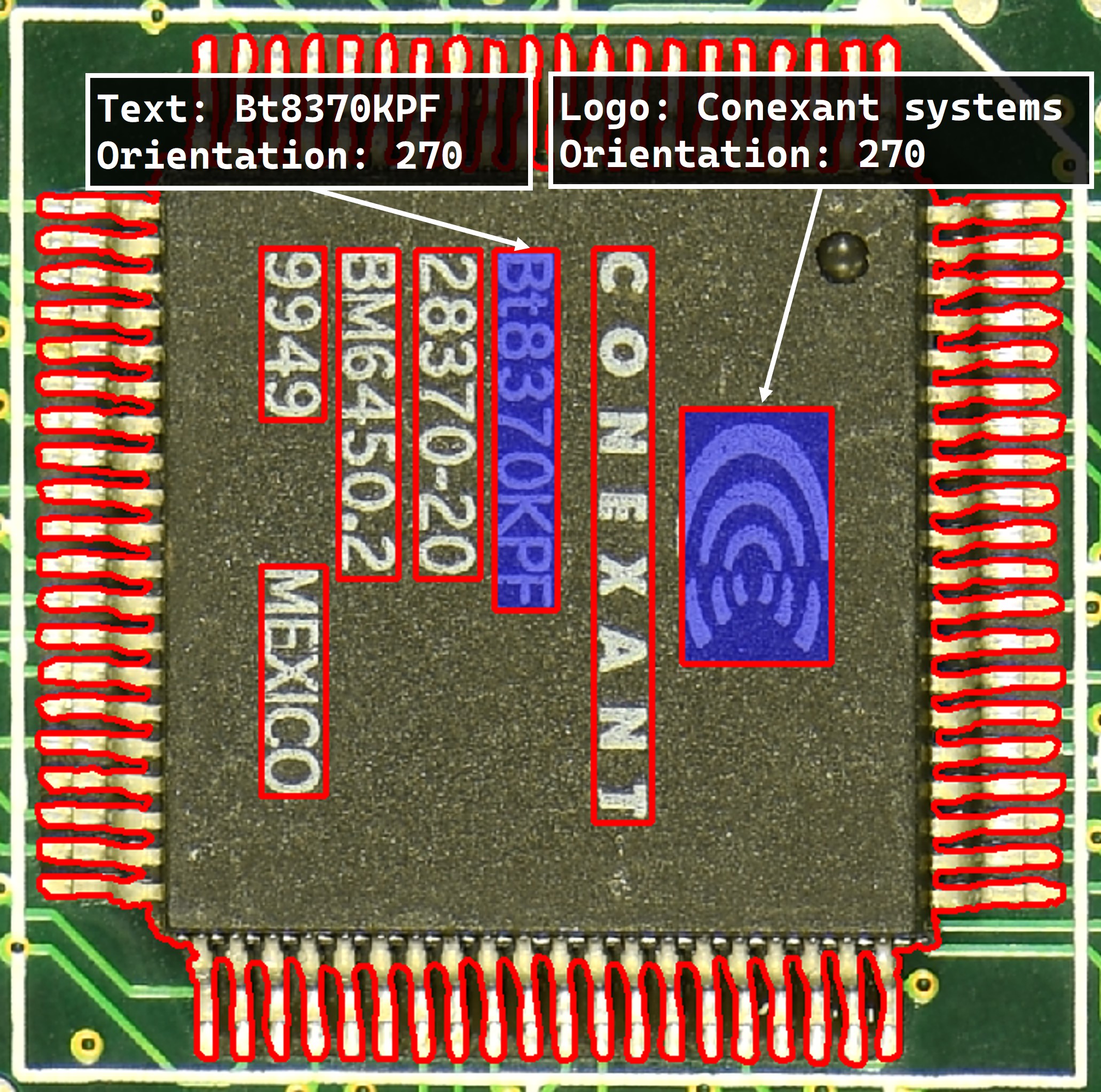}
	\caption{Representative text and \ac{smd} annotations. Precise contours, logo, text, and orientation information is captured along with additional metadata as described in Section \ref{sec:fpic}.}
	\label{fig:sample_annotated_component}
\end{figure}
}

\newcommand{\makePowerTraceFig}{%
\begin{figure}[!htbp]
	\centering
	\includegraphics[width=0.4\linewidth]{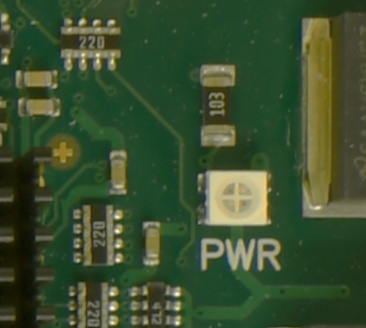}
	\caption{Traces for power lines require larger widths in general compared to normal signal traces to handle increased current flow. It can be helpful to verify this information against \ac{ocr} information.}
	\label{fig:power_trace}
\end{figure}
}

\newcommand{\makeOcrFieldsTable}{%
\begin{table}[!htbp]
    \centering
    \includegraphics[width=0.8\linewidth]{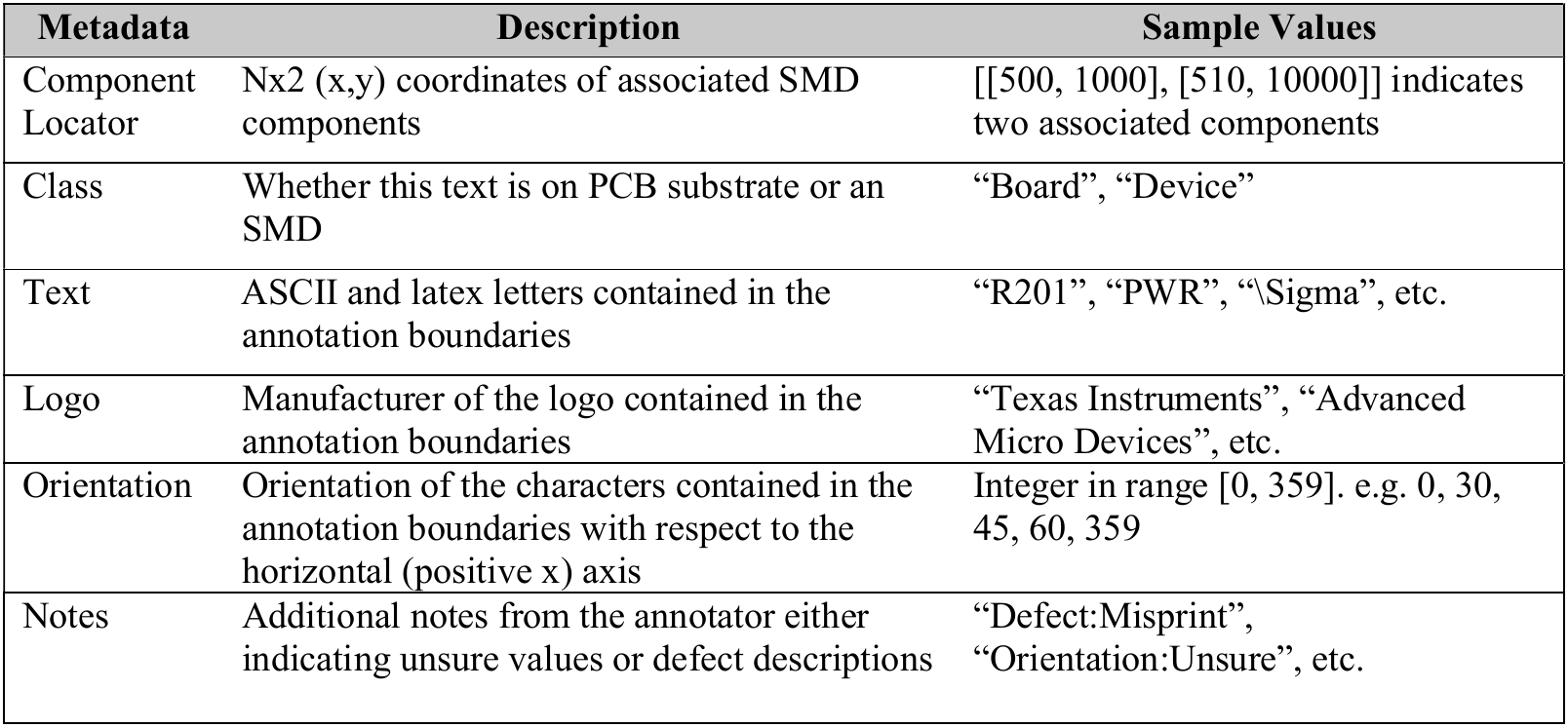}
    \caption{Metadata collected during text annotation}
    \label{tab:ocr_fields}
\end{table}
}

\newcommand{\makeMethodologicalChallengesTable}{%
\begin{table}[!tb]
    \centering
    \includegraphics[width=0.8\linewidth]{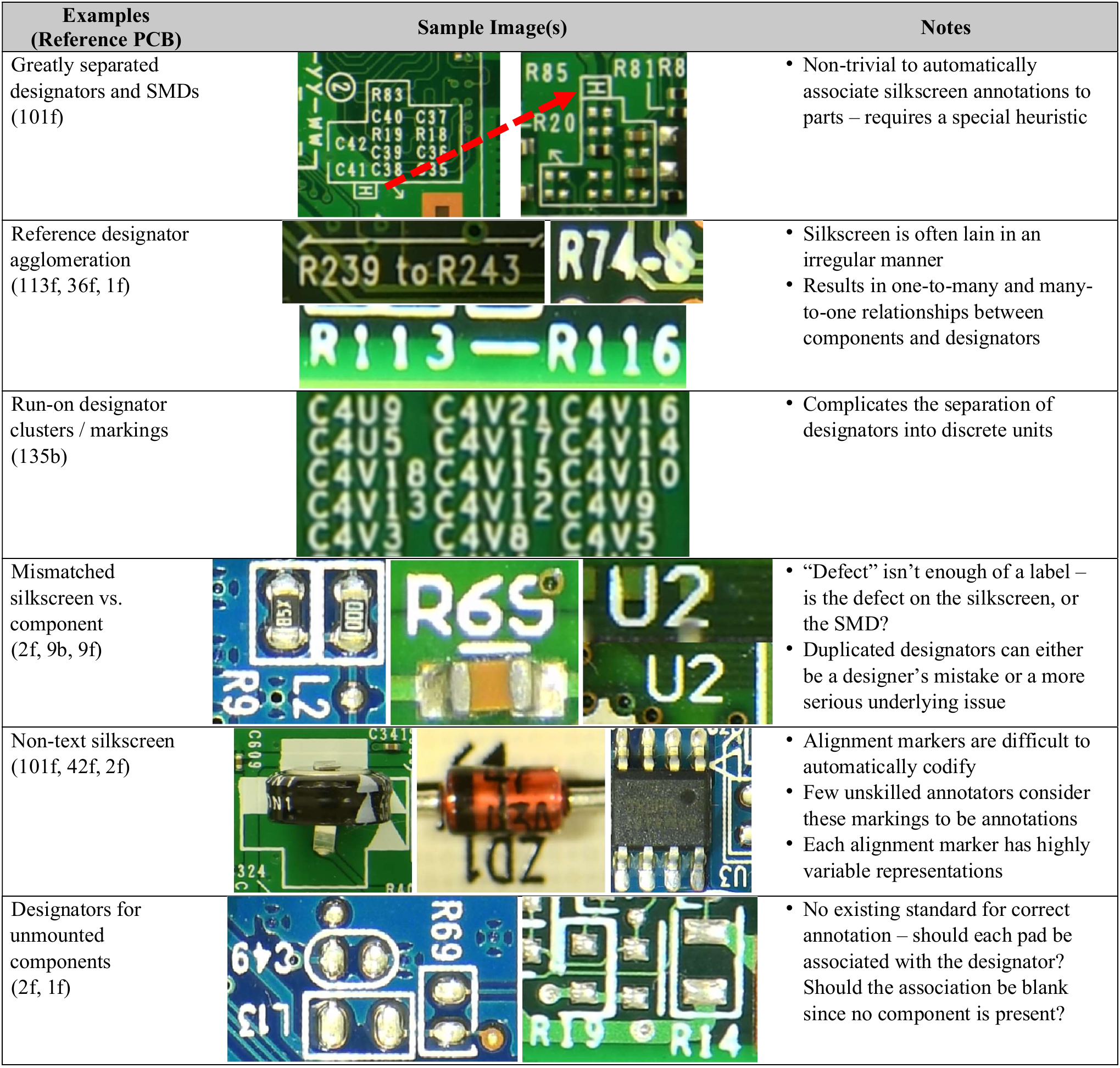}
    \caption{Overview of the methodological challenges encountered during data annotation. Regardless of image quality or inspection method, these issues will be present when creating a store of ground truth information.}
    \label{tab:methodological_challenges}
\end{table}
}

\newcommand{\makeImagingChallengesTable}{%
\begin{table}[!tb]
    \centering
    \includegraphics[width=0.8\linewidth]{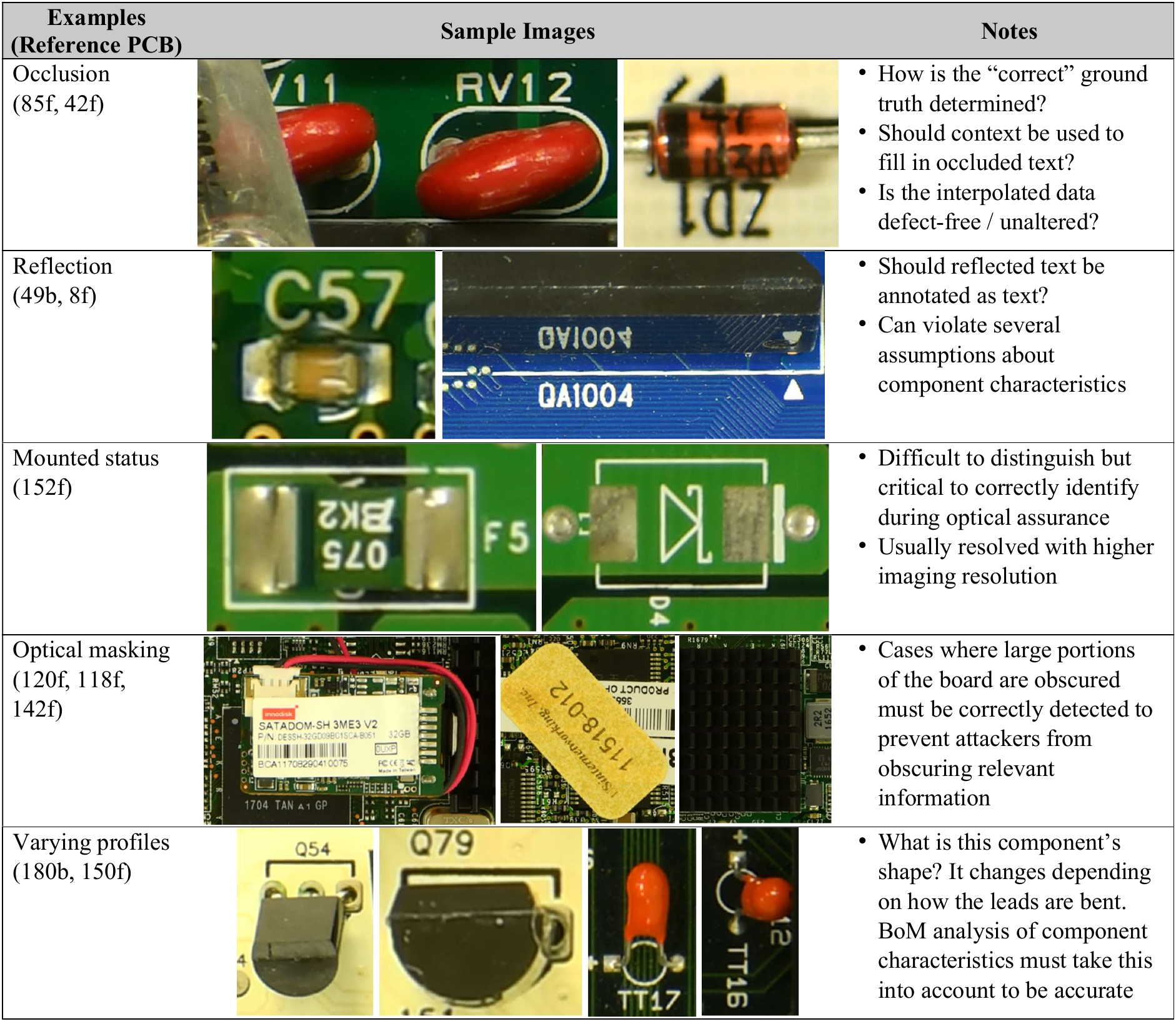}
    \caption{Overview of challenges faced by image acquisition parameters. In other words, it is possible to alleviate some of these concerns by altering various aspects of the imaging setup.}
    \label{tab:imaging_challenges}
\end{table}
}

\DeclareAcronym{pcba}{
  short=PCBA,
  long=PCB assembly,
  long-plural-form=PCB assemblies,
}
\DeclareAcronym{pcb}{
  short=PCB,
  long=printed circuit board,
}
\DeclareAcronym{smd}{
  short=SMD,
  long=surface-mount device,
}
\DeclareAcronym{hw}{
  short=HW,
  long=hardware,
}
\DeclareAcronym{ic}{
  short=IC,
  long=integrated circuit,
}
\DeclareAcronym{bom}{
  short=BoM,
  long=bill of materials,
}
\DeclareAcronym{sme}{
    short=SME,
    long=subject matter expert,
}
\DeclareAcronym{drc}{
    short=DRC,
    long=design rule check,
}
\DeclareAcronym{ipc}{
    short=IPC,
    long=Institute for Interconnecting and Packaging Electronic Circuits,
}
\DeclareAcronym{cad}{
    short=CAD,
    long=computer-aided design,
}
\DeclareAcronym{oem}{
    short=OEM,
    long=original equipment manufacturer,
}

\DeclareAcronym{aoi}{
  short=AOI,
  long=automated optical inspection,
}
\DeclareAcronym{ir}{
  short=IR,
  long=infrared,
}
\DeclareAcronym{thz}{
  short=THz,
  long=Terahertz,
}
\DeclareAcronym{ocr}{
  short=OCR,
  long=optical character recognition,
}
\DeclareAcronym{ml}{
  short=ML,
  long=machine learning,
}
\DeclareAcronym{ai}{
  short=AI,
  long=artificial intelligence,
}
\DeclareAcronym{cv}{
  short=CV,
  long=computer vision,
}
\DeclareAcronym{svm}{
  short=SVM,
  long=support vector machine,
}
\DeclareAcronym{fpic}{
  short=FPIC,
  long=FICS PCB Image Collection,
}
\DeclareAcronym{fics}{
  short=FICS,
  long=Florida Institute for Cybersecurity Research,
}

\DeclareAcronym{ppmm}{
  short=ppmm,
  long=pixels per millimeter,
}

\DeclareAcronym{ip}{
  short=IP,
  long=intellectual property,
}
\AtBeginDocument{%
	\providecommand\BibTeX{{%
			\normalfont B\kern-0.5em{\scshape i\kern-0.25em b}\kern-0.8em\TeX}}}

\setcopyright{acmcopyright}
\copyrightyear{2023}
\acmYear{2023}
\acmDOI{XXXXXXX.XXXXXXX}

\acmConference[JETC '23]{Make sure to enter the correct
	conference title from your rights confirmation email}{XXX XXX 2023}{XX, XX}
\acmPrice{15.00}
\acmISBN{978-1-4503-XXXX-X/18/06}

\begin{document}
	
	\title{FPIC: A Novel Semantic Dataset for Optical PCB Assurance}
	
	\author{Nathan Jessurun}
	\email{njessurun@ufl.edu}
	\author{Olivia P. Dizon-Paradis}
	\email{paradiso@ufl.edu}
	\author{Jacob Harrison}
	\email{jacob.harriso5@ufl.edu}
	\author{Shajib Ghosh}
	\email{shajib.ghosh@ufl.edu}
	\author{Mark M. Tehranipoor}
	\email{tehranipoor@ece.ufl.edu}
	\author{Damon L. Woodard}
	\email{dwoodard@ece.ufl.edu}
	\author{Navid Asadizanjani}
	\email{nasadi@ece.ufl.edu}
	
	\affiliation{%
		\institution{University of Florida}
		\streetaddress{Florida Institute for Cybersecurity Research}
		\city{Gainesville}
		\state{Florida}
		\country{USA}
		\postcode{32603}
	}
	
	\renewcommand{\shortauthors}{Jessurun et al.}
	
	\begin{abstract}
		Outsourced \ac{pcb} fabrication necessitates increased hardware assurance capabilities. Several assurance techniques based on \ac{aoi} have been proposed that leverage \ac{pcb} images acquired using digital cameras. We review state-of-the-art \ac{aoi} techniques and observe a strong, rapid trend toward \ac{ml} solutions. These require significant amounts of labeled ground truth data, which is lacking in the publicly available \ac{pcb} data space. We contribute the \ac{fpic} dataset to address this need. Additionally, we outline new hardware security methodologies enabled by our data set.
	\end{abstract}


\maketitle

\section{Introduction}
\Acp{pcb} are key components of many modern electronic systems, from computers and mobile phones in the private sector to military and medical devices in the public sector. As outsourced manufacturing becomes increasingly common, electronic systems are left vulnerable to supply chain threats like malicious modification, reverse engineering and IP theft, and counterfeiting~\cite{robertson_big_2018,appelbaum_nsas_2013, harrison_malicious_2021, senate_investigation_2012}. Subsequently, malicious entities with even minor supply chain access can severely compromise national security through Trojan insertions and related device modifications. Hence, techniques for auditing \ac{pcb} correctness and component authenticity are needed. Due to its favorable trade-offs between speed, accuracy, and cost compared to X-Ray and other wavelengths, optical inspection of \ac{pcb} components stands out as a promising inspection modality~\cite{moganti_automatic_1996}.

Since defects and Trojans are often highly complex, \ac{cv} solutions alone cannot provide robust system-level assurance. Hence, optical solutions are trending toward \ac{ai} and \ac{ml} methods for a generalizable, scalable approach. Outside of \ac{pcb} analysis, most AI/ML developments occur in domains with numerous large datasets such as medical imaging, aerial imaging, generic object recognition, etc. Unfortunately, the same is not true for optical \ac{pcb} assurance -- remarkably few datasets are publicly available, and each possesses a relatively small number of images compared to other domains~\cite{richter_development_2017}. Transfer learning seems promising for mitigating this lack of large \ac{pcb} data set, but it  requires similarities between the source and target datasets. \ac{pcb} data varies significantly from datasets like PascalVOC and ImageNet which are staples in other fields -- this complicates the use of transfer learning to bootstrap larger datasets. \emph{As a result, there is a critical lack of available \ac{pcb} data to train AI/ML solutions for robust \ac{pcb} assurance.}

We contribute the FICS\footnote{\ac{fics}} PCB Image Collection (FPIC) dataset to address this deficiency. First, \ac{fpic} contains more unique \ac{pcb} images with labeled instances than any prior work in the \ac{pcb} assurance domain. Further, it includes novel metadata such as semantic annotation boundaries and component-to-silkscreen correlations. Some representative annotations are shown in \autoref{fig:sample_annotated_component}. Finally, this paper outlines several ways the FPIC dataset provides a unique groundwork for advancing hardware assurance techniques. The dataset is available at \url{https://www.trust-hub.org/#/data/pcb-images}. Critically, this dataset will grow over time to continually expand its breadth of \ac{pcb} coverage. Regular updates are already planned for the next 12 months, with more contributions anticipated from the community once semi-automated data collection is fully enabled.

\makeSampleAnnotatedComponentFig

Section \ref{sec:background} discusses relevant threats to \acp{pcb}, argues that optical imaging is a promising solution, and explains the trend toward \ac{ml} solutions to this problem. Section~\ref{sec:fpic} discusses the newly collected \ac{fpic} dataset and fundamental optical \ac{pcb} assurance challenges encountered through its development.
Section~\ref{sec:method-insights} outlines methodological considerations resulting from the data collection procedure and hardware assurance ramifications.
In Section~\ref{sec:newly-enabled-research}, \ac{fpic}'s ramifications on state-of-the-art techniques are proposed, and how its data will propel the field of optical \ac{pcb} assurance forward. Finally, Section \ref{sec:conclusion} concludes with remarks about how the dataset will continue to grow and impact the field of \ac{aoi}.
\section{Background}\label{sec:background}
Most aspects of cost-effective \ac{pcba} assurance center around \ac{aoi} capabilities. The following subsections outline what threats are relevant, why this method is a popular approach toward a solution, and how \ac{aoi} approaches trend toward machine learning tactics over time. The section concludes with a description of the datasets enabling \ac{ml}-based \ac{aoi}.

\subsection{Threats to PCBAs} \label{sec:background-attacks-on-pcbs}
Optical assurance has been proposed for identifying counterfeit and maliciously-modified (``trojan'') components and \acp{pcba}. Automated optical inspection also has applications for \ac{pcba} reverse-engineering.

The term \textit{counterfeit} describes components and assemblies that are either a) misrepresented by suppliers as having different function, parameters, or performance or b) produced illegitimately, e.g., from stolen \ac{ip}.\footnote{See \cite{guin_counterfeit_2014} for a detailed counterfeit electronics taxonomy.} Counterfeit electronics siphon revenue from legitimate manufacturers and threaten the reliability and security of electronic systems~\cite{senate_investigation_2012, grow_dangerous_2008, guin_counterfeit_2014}.

\textit{Trojan hardware} contains intentional malicious modifications that compromise confidentiality of sensitive information, system integrity or performance, or cause denial of service. Trojan integrated circuits have been extensively investigated, but attacks at the board-level have only recently garnered attention. Unlike \acp{ic}, which are assumed vulnerable only during manufacturing due to challenges of \ac{ic} editing, \acp{pcba} are vulnerable to both manufacturing-time modifications (e.g., the electromigration modification proposed by~\cite{mcguire_pcb_2019}) and post-manufacturing attacks (e.g., video game modchips~\cite{huang_keeping_2002} and implants from the NSA ANT catalog~\cite{appelbaum_nsas_2013}).

Automated \ac{pcba} inspection has also been proposed to assist reverse engineering, e.g., by automatically extracting a \ac{pcba}'s \ac{bom}. Imaging-assisted reverse engineering could be used for good by helping repair or replace obsolete equipment or contribute to assurance, or for evil by cloning or pirating legitimate \acp{pcb}~\cite{quadir_survey_2016}.


\subsection{Automated optical inspection \& \ac{ai}/\ac{ml}}
While some methods for hardware assurance have existed since the beginning of the digital age, other techniques have only grown popular in recent years due to increased computing power and advancements in \ac{ai}. The sections below outline trends in the inspection process, and their relevance toward current inspection capabilities.

\subsubsection{Overview}
\ac{aoi} is one of the oldest and most common methods for \ac{pcb} assurance, with some of the earliest references in the 1960s \cite{watkins_inspection_1969,moganti_automatic_1996}. However, preliminary efforts were largely confined to pre-assembly inspection and photomask evaluation. As a result, aspects like \ac{smd} solder/placement inspection and counterfeit analysis were not considered. Furthermore, optical inspection was limited by several constraints including hard-coded specifications, limited-area inspection, and significant amounts of manual intervention \cite{moganti_automatic_1996,wang_machine_2017,hani_review_2012}. Though exponential improvements in computing power alleviated some of these concerns, \acp{sme} still played a prominent role in the process. As a result, despite several decades of published research on various aspects of \ac{aoi}, fully integrated \ac{pcba} inspection remained heavily unexplored. Fortunately, recent developments in machine learning toward image recognition, defect detection, segmentation, localization, and more in other domains have been successfully translated to hardware assurance purposes, bridging several of these gaps.

The relevance and drawbacks of these \ac{ai} advancements change depending on whether assurance is performed with known genuine reference samples, as highlighted below. However, both mechanisms share the goal of characterizing fundamental aspects of given regions of interest, such as component type, (i.e. \ac{ic}, resistor, capacitor, etc.) defect type (i.e. short, bridge, cavity, etc.), silkscreen / text information, etc. \cite{wei_cnnbased_2018,zhang_convolutional_2018,lu_ficspcb_2020,youn_automatic_2014,adibhatla_detecting_2018,lim_smd_2019,kim_smd_2018,huang_pcb_2019}.

\emph{Component Inspection}.
Several works have considered how optical inspection could be used to spot defects caused by harsh counterfeiting processes. For example, in recycling, which is believed to be a leading source of counterfeit parts~\cite{guin_a_comprehensive_2014}, workers heat \acp{pcba} to melt solder and then pry, yank, or beat components off of \acp{pcb}. Harvested parts may be washed in the river, laid on the ground to dry, and thrown haphazardly into bins~\cite{senate_investigation_2012}. Guin et al. enumerate physical defects from recycling and other counterfeiting processes~\cite{guin_a_comprehensive_2014}. In practice, subject matter experts search for these defects~\cite{senate_investigation_2012,goetz_counterfeit_2017} but manual inspection is unscalable; application of computer vision and machine learning to automatically find evidence of counterfeiting is a topic of active research. An alternative approach aims to uniquely identify components using optical inspection. For example, Wu et al. also use an automated approach to identify components beyond tolerance for lead placement, orientation, and more~\cite{wu_automated_2010}.

\emph{Assembly inspection}.
Rigorous detection of bogus \acp{pcba}~\cite{grow_dangerous_2008, tehranipoor_invasion_2017} has received relatively little attention. Generally, unsophisticated counterfeits can be spotted by discrepancies between genuine and suspect systems (e.g., \cite{noauthor_recognize_nodate, newman_anatomy_2020}). A few whole-\ac{pcba} automated inspection projects are under development~\cite{azhagan_new_2019} but are immature. X-ray analysis of bare \acp{pcb} has been proposed but cannot be applied to assemblies because mounted components cause x-ray artifacts~\cite{asadi_pcb_2017}. Often, optical \ac{pcba} assurance reduces to inspection of individual components. However, note that when the entire \ac{pcb} image is wholly evaluated, rather than regional subsets, there is no need to individually identify surface-mounted components. This occurs in quality assurance processes when the entire \ac{pcb} can be considered defective or non-defective rather than distinguishing integrity at the component level \cite{fridman_changechip_2021,ganapathy_defect_2021}.

\subsubsection{\ac{aoi} Using Golden Samples}
When a known authentic \ac{pcba} (a ``golden" sample) is available, defect and trojan analysis mainly consists of differential comparisons between the sample under test and the golden counterpart. Any change in optical characteristics between the two can be attributed either to environmental effects or component/board alterations. Because of its scalability, low cost, efficiency, and accuracy, this differential approach is the most popular form of \ac{aoi} \cite{moganti_automatic_1996}. These inspections can take place at both the component and assembly level, incorporating information from \ac{cad} schematics and other available domain knowledge. In most cases, a test sample is aligned against the golden reference, features are extracted, and a comparison or feature classification is performed between the golden and test samples. Resulting classification differences indicate deviations between golden and test samples. For instance, De Oliviera et al. discovered modifications made to fuel pump controller \acp{pcb} by identifying differences between reference and modified samples using SIFT features and a \ac{svm} classifier~\cite{mazondeoliveira_detecting_2017}. Zhao et al. used a machine vision system to locate known optical test points in an online fashion to perform sample inspection. Once the correct location is identified, feature extraction and pattern matching are performed as previously described~\cite{zhao_ni_2009}. Wang et al. analyzed the correlation coefficient between reference and test samples to identify scratches and other defects on \ac{ic} surfaces~\cite{wang_machine_2013}.

\subsubsection{\ac{aoi} without Golden Samples}
The benefits of golden sample-based \ac{aoi} are significant, but genuine \ac{pcba}s are often not available. When boards are past their supported life span (as is common for long-term government/military applications), highly customized, poorly documented, or of unknown origin, there is often little to no information available about the board, let alone golden images for differential image analysis. Alternative assurance methods must be employed to achieve necessary results. In contrast to golden-based \ac{aoi}, these methods must localize / characterize regions of interest and extrapolate relevant information in place of differential analysis. This is typically performed using a machine learning backbone and significant amounts of labeled data, as shown in \autoref{fig:aoi_pictoral_overview}.

\makeAoiPictoralOverviewFig

Component identification is responsible for analyzing previously acquired results and making an assurance assessment. During this process, elements of the physical sample are compared to ideal design standards to find likely evidence of manipulation or defects. In this manner, while the precise metrics for a genuine sample are not known beforehand, evidence of tampering will appear as regions which violate \acp{drc}, fall outside \ac{ipc} standards, do not correspond to a logical netlist/BoM extraction, or exhibit related deviations from expected behavior. In this vein, AutoBoM is a proposed framework which attempts to leverage this information, creating a tentative BoM from optically gleaned information and comparing it to known general component properties~\cite{azhagan_new_2019}. Lin et al. employ a coupled system of \ac{ocr} and image analysis to identify \acp{ic} with clear and blurry text in \ac{pcb} images~\cite{lin_using_2019}.

As described earlier, datasets play a critical role in the quality of machine learning-based assurance methods. Increasingly complex methods are developed for golden-less \ac{aoi}, but these techniques are heavily stunted without prior growth in the associated datasets. Thus, a discussion of currently available datasets provides reasonable insight into how robust current methods can be expected to perform and how much room for model architecture growth exists.

\subsection{Existing Datasets}\label{sec:background-existing-datasets}
Several publications introduce datasets useful for \ac{aoi}/\ac{ml} methods, each highlighting different aspects of optical \ac{pcb} inspection. While some focus on surface-mount devices, others provide information on connectivity such as traces and vias. The datasets are grouped into two categories below based on whether they are publicly accessible, ordered by the year they were introduced. These are summarized in addition to the proposed \ac{fpic} dataset in \autoref{tab:dataset_comparison}.

Notably, there are far fewer datasets for \ac{pcb} assurance than standard computer vision solutions, such as cell counting, aerial photos, and generic image segmentation \cite{pramerdorfer_dataset_2015}. This is a critical issue for a field like hardware assurance, since significant data is one of the few robust mechanisms to combat a dynamically evolving environment \cite{moganti_automatic_1996}. As of 2021, the largest publicly available datasets consist of at most 165 \ac{pcb} samples and at most 8,016 unique \ac{smd} annotations (PCB-DSLR and WACV respectively, see Section \ref{sec:fpic}). As a consequence, this increase in \ac{ml} applications for \ac{pcb} assurance without corresponding levels of data \textit{does not} equate to increased reliability and confidence metrics.
These concerns can be addressed either with smaller \ac{ml} architectures or larger amounts of ground truth data. While data augmentation can partially address this issue, a large set of \textit{unique, representative} data is essential for generalizable assurance, especially when tested on deep neural networks \cite{larochelle_exploring_2009}.

\subsubsection{Publicly Available}
\emph{CD-PCB (2021)} \cite{fridman_changechip_2021}. Fridman et al. provide 17 image pairs where each set contains images with and without synthetic defects. This dataset can provide insight into the general, board-level characteristics of defective samples.

\emph{Amazon lookout (2021)} \cite{ganapathy_defect_2021}. Ganapathy and Gupta demonstrate an AWS workflow to identify anomalous versus normal \acp{pcb} based on high-resolution optical images. Without instance-level labels, usability is constrained to high-level defect detection and general characteristic analysis. Nonetheless, they demonstrate reasonable accuracy in discriminating normal and anomalous samples with a given machine learning model.

\emph{IC-ChipNet (2020)} \cite{reza_icchipnet_2020}. Reza and Crandall present a curated bank of over 6,000 \ac{ic} images. While they do not provide general text annotations, each \ac{ic} is labeled with its manufacturer and several unique logos per manufacturer are represented. This dataset is useful for fine-grained retrieval, recognition, and verification of \ac{ic} images.

\emph{PCBExperiment (2020)} \cite{karanth_pcbexperiment_2020}. Not associated with a specific publication or method, this publicly available dataset consists of several hundred defective and non-defective \ac{pcb} samples. Its category depends on whether it passed a manual quality assurance check. Similar to Amazon Lookout above, no other annotations or instance-level attributes are defined in this dataset. However, as with CD-PCB it can be useful for generalizing traits of poor-quality boards or defective samples.

\emph{FICS-PCB (2020)} \cite{lu_ficspcb_2020}. Over 400 images were collected of 31 \ac{pcb} samples in this work. A combined total of 30,000 \ac{smd} bounding boxes are present across these duplicate images, with several pieces of metadata associated with each component. More information is visible in \autoref{tab:dataset_comparison}. The comprehensive data collected is useful for a variety of purposes, including resolution requirement analysis, component localization \& classification, and character recognition.

\emph{WACV (2019)} \cite{kuo_dataefficient_2019}. Kuo et al. provide one of the first public datasets with board text annotations. With over 6,000 component annotations and 10,000 text annotations across 41 \ac{pcb} images, they allow for a dynamic array of new machine learning tasks in addition to component detection and localization.

\emph{HRIPCB (2019)} \cite{huang_pcb_2019}. Huang and Wei present this dataset containing 10 \ac{pcb} images with multiple synthetic defects. In each case, bounding box annotations around each defect location are provided. Compared to \cite{karanth_pcbexperiment_2020,fridman_changechip_2021}, these annotations provide more precise and detailed information about defect compositions.

\emph{PCB-DSLR (2015)}
\cite{pramerdorfer_dataset_2015}. One of the first public \ac{pcb} datasets, Pramerdorfer and Kampel paved the way for machine learning applications with this contribution. The dataset contains 165 unique \acp{pcb} and over 2,000 labeled \ac{ic} instances with text data for a small subset. More images are captured in different orientations, increasing the size of data when label repetitions are considered.

\subsubsection{Out-of-Scope}
Several additional works highlight datasets that either are not publicly available or do not directly handle \ac{pcb}/\ac{smd} images.
\cite{gang_coresets_2020} presents a sizable dataset of \ac{pcb} images at a production plant and labels ``coresets'' of \ac{ocr} characters, which are building blocks for more flexible text-based model training. No public link is given, but the authors note in \cite{gang_character_2021} that data is available upon request.
DEEP-PCB (2019) provides a dataset of annotated substrate defects. Images consist of post-thresholding via/trace masks rather than the raw optical data \cite{tang_online_2019}.
A dataset of google images seeded by PCB-DSLR is presented by Chen et al. \cite{chen_addressing_2017}, but this curated and annotated list of images is not publicly available.
\cite{li_multisensor_2021} collects multimodal data including \ac{ir} and visible images of \ac{pcb} defects, but is not publicly accessible.
\cite{shieh_applying_2021} demonstrates the performance of a new YOLO (v5) architecture specifically designed to locate tiny defects in high-resolution quality-control \ac{pcb} images.
PCB-METAL presents a large dataset of high-resolution \ac{pcb} images similar to FICS-PCB, but there is no public link to the data and fewer representative component types are labeled \cite{mahalingam_pcbmetal_2019}.
\section{FPIC Dataset}\label{sec:fpic}
As noted previously, the \ac{fpic} dataset was developed to address data availability bottlenecks in \ac{ml}-based \ac{aoi} methods. The dataset consists of 261 images of 93 separate \acp{pcb}. Note that more images exist than physical \ac{pcb} samples since both front and back images can be acquired, and some \acp{pcb} are photographed in multiple settings. Both text and mounted components are annotated where applicable in most images, resulting in over 71,000 annotated instances. \ac{pcb} samples were purchased online or disassembled from a variety of different devices, e.g. computer hard drive controllers, servers, and audio amplifiers. An example image with annotations is shown in \autoref{fig:annotated_image}. \ac{fpic}'s creation and the rationale for included metadata is explained in the following subsections. Critically, this dataset will continue to grow larger over time as more samples are cataloged and annotated.

\makeAnnotatedImageFig

\ac{fpic}'s dataset statistics in comparison to other works is shown in \autoref{tab:dataset_comparison}, though fairly  comparing FPIC to  prior datasets is deceptively challenging. Comparing quantity of ground truth annotations between datasets requires that a distinction be made between \textit{unique} and \textit{total} instances: one \ac{pcb} might be photographed and annotated four times, but only one set of those annotations provide new information because the remaining three images are simply geometric transforms of the first. Machine learning algorithms must be careful to distinguish between these when using training sets, otherwise they risk treating augmentations as unseen samples. In an effort to compare datasets as fairly as possible, \autoref{tab:dataset_comparison} lists both the total and unique annotation quantities for datasets with duplicate annotations.

\makeDatasetComparisonTable

\subsection{Imaging}
Images were taken in controlled conditions with a Nikon D850 DSLR camera. A two-second delay mitigated camera shake during acquisition, and a two-second exposure reduced image noise. Each time the camera position or zoom changed, a calibration photo of an X-Rite Passport or Nano Color Checker was collected to normalize colors across changing environmental conditions (e.g., photographs taken with different amounts of natural light).
A portion of the dataset consists of the same \ac{pcb} samples imaged under several illumination and camera sensor parameters to study the effects of normalization \cite{paradis_color_2020}.

Each image in the database consists of a single photograph rather than a stitched sequence of tiles. While image stitching is common to increase resolution, it introduces defects  such as distortion / warping, imperfect alignment, and noise at stitching boundaries. A thorough investigation of these defects can be found in \cite{lu_ficspcb_2020}. \ac{fpic} optimizes for minimal processing and noise reduction at the expense of lower resolution for very large samples.

\subsection{Contour Annotation}
Collected images were processed using S3A (\url{https://gitlab.com/s3a/s3a}) and SuperAnnotate (\url{www.superannotate.com}). The SuperAnnotate service was responsible for creating high-fidelity semantic outlines around components in a subset of \ac{pcb} images as well as creating bounding boxes around device and board text. A significant number of images were annotated in-house using S3A, where assistance algorithms such as GrabCut and K-Means segmentation simplified this process. All contours (bounding box or otherwise) are represented as lists of polygons indicating either holes or foreground areas.

\subsection{Metadata Annotation}
Once gathered, component and text boundaries were loaded into S3A where the remaining metadata (Text, Logo, Designator,  etc.) was added locally. 
Since complex \ac{smd} contours were extraordinarily time-consuming to obtain, the only additional metadata associated with these instances was the ``Designator": silkscreen board text containing the component identifier such as `C', `L', `R', etc. If no designator was present, instead of attempting to infer the proper label, we opted not to assign a label to avoid adding potentially incorrect component labels. \autoref{fig:similar_looking_smds} illustrates one scenario where similar-looking components could easily be misidentified if annotators tried to guess component identity instead of relying only on designators.

\makeSimilarLookingSmdsFig

\subsection{Validation}
Each file underwent two rounds of manual validation and a logical check to minimize the number of human errors present in the final version. The logical check programmatically searched for edge cases such as replicated components in the same file, duplicate metadata, strongly overlapping regions, and other factors usually indicative of errors during annotation.


Unlike \ac{smd} components, \ac{ocr} instances have a host of associated metadata, including some fields that are present for the first time in \ac{pcb} dataset literature. These are summarized in \autoref{tab:ocr_fields}.

\makeOcrFieldsTable

\subsection{Database insertion}
The \ac{fpic} dataset is stored on TrustHub; \autoref{tab:dataset_folders} summarizes the database's directory structure. Beyond metadata for labeled instances, each image is associated with scale information to enable measurement of physical dimensions based on pixel counts, a color checker, and a camera model. The first 150 images in \ac{fpic} use the same \ac{pcb} samples at different scales to evaluate the effects of resolution on component detectability as done in \cite{lu_ficspcb_2020}.

Beyond color and scale calibration parameters, over three-fourths of the \ac{pcb} samples also possess a brief functional description, i.e. ``Honeywell 14500144-001 Module PLC PCB Board". This information enables searching for documents like data sheets or \ac{cad} files if desired.

\makeDatasetFoldersTable
\section{Methodological insights}\label{sec:method-insights}
After years of data collection, we learned important lessons as a result of sample diversity, variance between annotators, and acquisition volume. The metadata collected by \ac{fpic} (namely, semantic contours and silkscreen-to-component association) that was not included in prior work also yielded useful insights. These are explained in the following paragraphs, separated into imaging and logistical categories. Each subsection provides a handful of recommendations for improving future data collection.

\subsection{Imaging challenges}\label{sec:imaging-challenges}
Imaging challenges include imaging irregularities and particularities that can be resolved by altering the image acquisition setup. The examples below illustrate cases where artifacts of the imaging process alter ground truth markings depending on the details present in annotation instructions and annotator skill.

\makeImagingChallengesTable

\emph{Occluded components and markings}.
Some, but not all, occluded components and markings can be revealed by slightly adjusting the angle from the sample to the camera. Many markings that would be valuable for assurance, e.g., polarity markers, component symbols, or designators, are located directly beneath mounted components to assist in manual assembly or rework. Unfortunately, it is usually impossible to completely image these markings even when multiple camera angles are considered.The top row of \autoref{tab:imaging_challenges} illustrates this challenge. When considering occluded components, the output segmentation often varies across annotators since some may consider only the visible portion of a device while others will extrapolate its position.

\emph{Reflective components and conformal coating}.
Reflections also negatively impact the ease of annotation and feature detection. Conformal coating is typically applied to \acp{pcba} requiring environmental hardening, resulting in a sheen across short components. This effect is stronger with direct lighting. As a result, it can be difficult to determine the true color or shape of the covered device, leading to misclassification or poor boundaries. Additionally, direct lighting across tall components can reflect bright areas of the board resulting in artifacts like the right side of row 2. Note that annotation instructions do not often outline a procedure for these cases, so artifacts may appear as ground truth annotations when working with low-cost or minimal QA annotation pipelines. However, applying a polarized filter to the camera can greatly reduce these specular side effects. If more control can be gained over the imaging environment, these effects can also be mitigated through precise lighting or stereophotographic acquisition.

\emph{Ambiguity between mounted/unmounted pads}.
Components that share colors with the PCB substrate complicate the distinction between populated vs. unpopulated solder pads. As a result, annotators can mistake the two and increase the difficulty in training a neural network to locate \acp{smd}. Row 3 illustrates the difference between a green SMD and silkscreen between empty pads. Note that the depicted example is straightforward to resolve due to the high-resolution image, but a low-resolution counterpart greatly increases the difficulty. Two simple methods exist for increasing resolution: adjusting the zoom or exposure time of the image. While increasing the zoom is often preferable, it can lead to more artifacts if image stitching is required as a result. Alternatively, increasing exposure time cuts down on some forms of sensor noise but cannot resolve all forms of resolution-based complexity.

\emph{Masked PCB areas}.
While the first category addressed individually occluded components, another frequent scenario involves large swaths of a PCB covered by various objects like heat sinks, stickers, daughterboards, etc. Depending on the threat model, this eases a malicious actor's difficulty inserting untrusted components since they can be trivially covered as seen in row 4. Thus, robust inspection mechanisms must determine a way to identify areas of the board likely to contain components masked by a secondary object. A minor amount of sample preparation can avoid several of these concerns. By removing stickers and disconnecting daughter boards before imaging,  large amounts of previously obscured circuitry will become visible. When removal is not possible (e.g. heat sinks and daughter boards which must remain attached), subsurface imaging such as X-Ray can be incorporated. 

\emph{Widely varying component profiles}.
When components have long leads (i.e. transistors, diodes, crystal oscillators, etc.), they can be contorted into a variety of shapes as viewed from the top down. In these cases, similar to occluded component scenarios, width/length annotations from the top-down object view do not correspond to width/length values from a BoM reference. Annotating additional metadata, such as the location of leads in the image, can assist in determining whether component height, width, or length should be matched to the visible dimensions in the optical image.

\subsection{Logistical insights}\label{sec:methodological-challenges}
In contrast to imaging challenges, the following discussion about methodological logistics does not depend on image quality or other environmental factors. Rather, these are fundamental characteristics of an annotation and assurance workflow that attempts to use flowchart rules to encode human design insights. Each heading illustrates the difficulties of correlating silkscreen information with PCB design characteristics.

\makeMethodologicalChallengesTable

\emph{Greatly separated designators and SMDs}.
Ideally, designators are close to their associated components and unlikely to be confused with neighboring designators. However, this is not the case in densely populated PCB regions where component real estate imposes stringent requirements. In these scenarios, several designators are grouped together, placed in an open area on the board, and `linked' in some manner to a corresponding group of components as shown in \autoref{tab:methodological_challenges} row 1. The way in which links are formed depends entirely on the PCB designer; cluster labels, arrows, and orientation matching are common approaches, but there are others. These common approaches enable development of heuristics that may be able to associate designators with components in the common case, but as long as designator placement remains a stylistic decision it will be difficult to confidently automate associations.

\emph{Reference designator agglomeration}.
When several designators are clustered as previously described and all refer to a sequential group of similar component types (i.e. a row of resistors/capacitors), one approach is to create a single label for all components. Various manifestations of this principle are illustrated in row 2. Since there are no design standards for how this grouping must take place, where silkscreen must be located relative to components, or how the association is made between a component and its text, understanding and annotating this association is difficult. Additionally, these cases demonstrate how silkscreen-to-component associations are not a one-to-one mapping: a single silkscreen can refer to multiple components and multiple silkscreens can belong to the same component (this is fairly common in IC or header pin annotations).


\emph{Run-on designator clusters}.
The same space restrictions explained in the first paragraph can cause groups of designators to run into each other and appear as one long text string. Humans can usually intuit the designer's intent, but these run-on designators can complicate automatic component designator extraction using off-the-shelf OCR methods. In the example shown in row 3, not only are designators prefixed with an integer character (``C\ul{4}V"), but are close enough to nearby designators that whitespace cannot be used as token separator.

\emph{Mismatched silkscreen vs. component}.
From an assurance perspective, this is one of the most challenging scenarios to address. In prior works, independent silkscreen and component annotations could both be considered valid as shown in the first two figures of row 4. However, the component in question clearly does not match its reference designator. ``L" is most commonly reserved for inductors, while ``R" is almost exclusively used for resistors. From their appearance, the associated \acp{smd} are clearly a resistor and capacitor, respectively. As such, while the components are properly mounted and the text is legible, there is an error in their association. Similarly, the third figure in this row illustrates a duplicate silkscreen marking. Both are correct for their associated component and legible, but the duplicate reference is highly abnormal and likely indicates a silkscreen mistake. In cases like these, determining whether the error is with the design, silkscreen print, or mounted component creates challenges for assurance.

\emph{Non-text silkscreen}.
Beyond alphanumeric characters, a host of various additional information is printed on the PCB substrate. As described in Section \ref{sec:imaging-challenges}, silkscreen can also refer to mounting information. However, a common theme is that silkscreen is primarily intended for human evaluation and has no uniform representation. Hence, without a catalog of common templates, it can be complex to algorithmically derive alignment information as presented in row 5 of \autoref{tab:methodological_challenges}. In the first instance, the polarity of a capacitor is presented with double arrows that should match the on-device markings present. Secondly, the obscured Zener diode symbol indicates polarity with a vertical line that should match the black bar present on the mounted component. Finally, the most complex case involves overlapping silkscreens where a left-facing triangle indicates the position of an IC's first pin. In each case, a vastly different visual cue is used to indicate alignment. These complexities drastically complicate how features such as polarity should be represented in an annotation database.

\emph{Designators for unmounted components}.
The final highlighted case for methodological insights involves designator associations with unpopulated solder pads. Technically, the designator does not refer to the pads, but the missing component meant to reside between them. At the same time, the pads are a reasonable approximation of the same spatial location associated with the text. Hence, there is justifiable reasoning for both including and excluding a correlation indicator for the silkscreen in question. Moreover, if a component association is created, it is unclear whether it belongs to one or multiple pads. Each of these considerations complicates the process of generating ground truth labels in this circumstance. during FPIC annotation, the pad closest to the designator is given the association.


\section{Newly enabled research}\label{sec:newly-enabled-research}




\ac{fpic} is larger and contains more diverse \ac{pcb} examples than prior \ac{pcb} image datasets, making it a better input to \ac{ml} training and a better benchmark for validating optical assurance techniques. Additionally, its annotations include novel information that enables new optical assurance research. The following paragraphs elucidate the connections between specific \ac{pcb} assurance challenges and data that is collected for the first time in \ac{fpic}.


\subsection{Dataset size and diversity}

The left-hand side of \autoref{fig:aoi_pictoral_overview} emphasizes that optical assurance must be calibrated for the \ac{pcb} under inspection and the threat to be detected. \acp{pcb} exhibit tremendous diversity across many characteristics that could impact optical assurance performance. For example, densely-packed boards, smaller components, or more sophisticated counterfeiting might all require higher imaging resolution to enable successful assurance. If a dataset is systematically biased with regard to imaging-relevant characteristics (e.g., a dataset containing boards of a similar manufacturing year, similar application domain, similar types of components, etc.), this may bias \ac{ml} networks trained on the data and limit the generality of performance claims that can be made for techniques that are validated against the dataset. Increasing the number and variety of represented \acp{pcb} and annotated components is the best hedge against these pitfalls.

As summarized in \autoref{tab:dataset_comparison}, \ac{fpic} is significantly larger and more generalized than in prior work.
It possesses more labeled \ac{pcb} samples and drastically more unique component annotations than the leading prior work. Among several other significant benefits, we highlight that a larger dataset allows tighter confidence margins on statistical analysis, enables \ac{fpic} to display more optical inspection corner cases, and provides more data for training large \ac{ai} networks. Notably, \ac{fpic} will continue to grow over time as more \acp{pcb} and annotations are added. As the technology landscape evolves, it is necessary to re-evaluate whether the dataset remains representative of boards to-be-assured.


While dataset size is important, it cannot stand alone as a metric of dataset quality; diversity of examples in the dataset is equally important. 
\ac{fpic} was purposefully built to include \acp{pcb} from many application domains, built by different \acp{oem}, manufactured in a range of years, and obtained from a variety of distributors in a variety of conditions. This intentional variation ensures that \ac{fpic} is representative of the broader population of printed circuit boards. This will help \ac{ai} models trained on \ac{fpic} to perform well on systems they have never seen and reduces bias when \ac{fpic} is used to validate optical assurance techniques.
Manufacturing date, \ac{pcb} function, manufacturer, distributor, etc. are qualitative and imprecise proxies for dataset diversity. Currently, there are no precise, measurable characteristics for describing a population of \acp{pcb} from an imaging perspective, but additional research will be able to standardize such metrics. Here, too, \ac{fpic} can provide insight. Its size and diversity enable studies to determine the most imaging-relevant \ac{pcb} characteristics.

\subsection{Improved SMD contour analysis from semantic segmentation}


Overwhelming amounts of literature demonstrate the usefulness of semantically segmented ground truth for enhanced machine learning capabilities \cite{garcia-garcia_review_2017,wang_understanding_2018,zhang_context_2018,long_fully_2015,yu_bisenet_2018}. While segmentation is possible with automated methods, the outputs are not consistently defect-free without significant hyperparameter tuning. In other words, high-quality automated segmentation still requires a large amount of human oversight in the segmentation architecture and parametrization. However, neural networks trained on human-verified segmentation masks overcome these difficulties with enough diverse ground truth data. The paragraphs below detail several ways machine learning tasks trained using segmentation ground truth rather than bounding boxes yield significantly improved outputs.

\emph{Component localization}.
While bounding box data can reasonably train a network to find components on a PCB, semantically trained networks can take this one step further. \autoref{fig:bbox_vs_semantic} shows that predictions overlap for nearby components in a bounding-box-trained LinkNet architecture versus its semantic counterpart. All aspects of training were the same in both cases except the masks.

\makeBboxVsSemanticFig

\emph{BoM property extraction}.
Multiple component footprint properties such as pin count, pitch, width, and spacing can only be determined with accurate outlines. FPIC provides enough of these samples to allow property extraction on unseen data as well by training semantic segmentation networks on the ground truth information. This drastically improves the ability of assurance algorithms to cross-reference known component properties against IPC standards and datasheet specifications. \autoref{fig:bbox_vs_semantic} also shows how a semantically trained ML model accurately represents IC contours with sufficient training data. \autoref{fig:bom_property_extraction} illustrates how these contours can be directly translated into BoM properties.

\makeBomPropertyExtractionFig

\emph{Assembly analysis}.
Beyond the component itself, how a device is mounted on the PCB can also be analyzed more fully with semantic contours. The difference between an oblong and right-angled device can yield insights as to whether the mount is within tolerance or is defective. Similarly, heavily skewed pins from tall through-hold components can be readily identified with precise boundaries and would otherwise not stand out with traditional bounding box annotations.

\subsection{Schematic analysis from component locator association}
Often, silkscreen designator information is critical to identifying the function or type of component present in an optical image. As such, identifying which components are associated with what board text can result in a significant boost in identification capabilities.

\emph{DRC Analysis}.
Text such as ``PWR" or ``GND" can be associated with the substrate itself rather than / along with a component. In these cases, aspects of the board layout can be analyzed in light of this silkscreen information to ensure both are in agreement. \autoref{fig:power_trace} demonstrates how the larger width requirement for power traces can be evaluated in light of the associated silkscreen identifier.

\makePowerTraceFig

\emph{Designator Verification}.
Datasheets and BoMs often make heavy use of pin numbers and reference designators (i.e. ``R101", ``L2", etc.). Since the association between components and their text can be highly nontrivial (see Section \ref{sec:methodological-challenges}), ground truth markers for this information are highly valuable for automated inspection efforts. When the designator is correct, and the SMD looks good, only the component locator can indicate whether there is a mismatch.

\emph{Assembly verification}.
As noted in \autoref{tab:methodological_challenges}, several types of silkscreen identifiers assist in determining device orientation or polarization. Hence, additional research objectives can bring an orthogonal dimension of assurance inspection to the traditional optical image analysis workflow.

Chatterjee et al. describe an augmented reality system that quickly correlates design schematics with hardware to simplify the process of quality assurance and HW/SW context switching \cite{chatterjee_augmented_2021}. Data from FPIC would allow such a system to rely less on software schematics for similar offerings when few design files are present.

\subsection{Data balance through parametrized acquisition}
Beyond the annotated metadata, much information about the PCB samples themselves is also collected. By combining information about real-world image scale, PCB manufacturer, and board description, SMD information can be correlated against additional factors other than their optical appearance. For instance, novel research directions might include evaluating whether certain manufacturers prefer specific brands or ratings of passive components. Or, whether defects and annotation errors occur more frequently for some sample categories. Significantly, the descriptive information associated with each board can be tied to relevant datasheets and CAD information for multiple samples, allowing schematic-level verification of collected information.
\section{Conclusion and Future Work}\label{sec:conclusion}


In conclusion, we have reviewed state-of-the-art techniques for \ac{aoi} and observed the strong, rapid trend toward \ac{ml} solutions. These require significant amounts of labeled ground truth data, which is lacking in the publicly available \ac{pcb} data space. The \ac{fpic} dataset (\url{https://www.trust-hub.org/#/data/pcb-images}) is proposed to address this bottleneck in available large-volume, diverse annotations. Additionally, this work covers the potential increase in hardware security capabilities and observed methodological distinctions highlighted during data collection.

\emph{Via, Trace, Pin Annotation}.
Future releases of the \ac{fpic} dataset will include annotations of vias, traces, and pins across a board's surface. This will yield significant insights into cross-component connectivity, the relationship between components and substrates, signal frequency analysis through trace layout inspection, and much more.

\emph{Synthesized \ac{pcb} samples}.
Beyond annotating purchased physical samples, future iterations can include fully synthesized \ac{cad} renderings and compare digital versus real-world annotation results. In this manner, the \ac{fpic} dataset could be drastically increased by providing augmented virtual counterparts. Moreover, a subset of these samples could be fabricated to determine more accurate relationships between design schematics and the acquired optical data. In this vein, Calzada et al. demonstrated several challenges inherent in optical \ac{pcb} inspection by designing custom samples with various intentional defects and Trojans, ranging from easy to difficult detection criteria~\cite{paul_gomac}.

\emph{Track related \ac{cad} schematics and datasheets}.
Along with physical samples, the item description is associated with datasheets or development files in several cases. Collecting this information in addition to each board will greatly increase the ability to provide \ac{hw} assurance through cross-referencing against known circuit properties. Additionally, these files would increase the accuracy of ground truth annotations since they provide references for the type and characteristics of each mounted component.

\emph{Multimodal data fusion}.
The \ac{fpic} dataset is a valuable resource for \ac{aoi} research, but does little to address volumetric issues, material properties, etc. Toward this end, increased data acquisition \emph{and rigorous labeling} in other modalities such as X-ray, \ac{thz}, and similarly neglected modalities would greatly assist additional assurance objectives. Active efforts in this direction are underway by the \ac{fics} group~\cite{mehta_fics_2022}, but as explained below it will be essential for additional collaborators to join this process.

\emph{Continuous collection and collaboration}.
A major theme throughout this work is the importance of publicly available data from a diverse array of sources and annotators. Continuous collaboration with the hardware assurance and computer vision communities at large is essential for growth in \ac{aoi} capabilities. More than just data alone, these enhancements would include code libraries, frameworks, standards, and evaluation mechanisms/metrics for improving the baseline for quality ground truth annotations. The \ac{fpic} dataset will continue to grow in the near future, and this increased coordination and collaboration will ensure the quality and fidelity of the dataset only improves in that course.

\printbibliography

@article{adibhatla_detecting_2018,
  title = {Detecting {{Defects}} in {{PCB}} Using {{Deep Learning}} via {{Convolution Neural Networks}}},
  author = {Adibhatla, Venkat Anil and Shieh, J. and Abbod, M. and Chih, Huan-Chuang and Hsu, C. and Cheng, Joseph},
  date = {2018},
  journaltitle = {2018 13th International Microsystems, Packaging, Assembly and Circuits Technology Conference (IMPACT)},
  doi = {10.1109/IMPACT.2018.8625828}
}

@misc{huang_keeping_2002,
	title = {Keeping secrets in hardware: the {Microsoft} {XBox} case study},
	publisher = {MIT AI Lab},
	author = {Huang, Andrew},
	month = may,
	year = {2002},
	note = {AI Memo 2002-08}}

@article{chatterjee_augmented_2021,
  title = {Augmented {{Silkscreen}}: {{Designing AR Interactions}} for {{Debugging Printed Circuit Boards}}},
  shorttitle = {Augmented {{Silkscreen}}},
  author = {Chatterjee, Ishan and Khvan, Olga and Pforte, Tadeusz and Li, Richard and Patel, Shwetak N.},
  date = {2021},
  journaltitle = {Conference on Designing Interactive Systems},
  doi = {10.1145/3461778.3462091}
}

@inproceedings{hani_review_2012,
  title = {A Review of {{SMD}}-{{PCB}} Defects and Detection Algorithms},
  booktitle = {Other {{Conferences}}},
  author = {Hani, A. F. Mohd and Malik, A. and Kamil, Raja and Thong, C.},
  date = {2012},
  doi = {10.1117/12.920531}
}

@inproceedings{kim_smd_2018,
  title = {{{SMD Defect Classification}} by {{Convolution Neural Network}} and {{PCB Image Transform}}},
  booktitle = {2018 {{IEEE}} 3rd {{International Conference}} on {{Computing}}, {{Communication}} and {{Security}} ({{ICCCS}})},
  author = {Kim, Young-Gyu and Lim, Dae-Ui and Ryu, Jong-Hyun and Park, Tae-Hyoung},
  date = {2018-10},
  pages = {180--183},
  doi = {10.1109/CCCS.2018.8586818},
  eventtitle = {2018 {{IEEE}} 3rd {{International Conference}} on {{Computing}}, {{Communication}} and {{Security}} ({{ICCCS}})}
}

@inproceedings{kuo_dataefficient_2019,
    title = {Data-Efficient Graph Embedding Learning for PCB Component Detection},
    author = {Kuo, Chia-Wen and Ashmore, Jacob and Huggins, David and Kira, Zsolt},
    booktitle={2019 IEEE Winter Conference on Applications of Computer Vision (WACV)},
    year = {2019},
    organization={IEEE}
}

@inproceedings{lim_smd_2019,
  title = {{{SMD Classification}} for {{Automated Optical Inspection Machine Using Convolution Neural Network}}},
  booktitle = {2019 {{Third IEEE International Conference}} on {{Robotic Computing}} ({{IRC}})},
  author = {Lim, Dae-ui and Kim, Young-Gyu and Park, Tae-Hyoung},
  date = {2019-02},
  pages = {395--398},
  doi = {10.1109/IRC.2019.00072},
  eventtitle = {2019 {{Third IEEE International Conference}} on {{Robotic Computing}} ({{IRC}})}
}

@article{quadir_survey_2016,
	title = {A Survey on Chip to System Reverse Engineering},
	volume = {13},
	issn = {1550-4832},
	url = {https://doi.org/10.1145/2755563},
	doi = {10.1145/2755563},
	number = {1},
	journal = {J. Emerg. Technol. Comput. Syst.},
	author = {Quadir, Shahed E. and Chen, Junlin and Forte, Domenic and Asadizanjani, Navid and Shahbazmohamadi, Sina and Wang, Lei and Chandy, John and Tehranipoor, Mark},
	month = apr,
	year = {2016},}

@article{moganti_automatic_1996,
  title = {Automatic {{PCB Inspection Algorithms}}: {{A Survey}}},
  shorttitle = {Automatic {{PCB Inspection Algorithms}}},
  author = {Moganti, Madhav and Ercal, Fikret and Dagli, Cihan H. and Tsunekawa, Shou},
  date = {1996-03-01},
  journaltitle = {Computer Vision and Image Understanding},
  shortjournal = {Computer Vision and Image Understanding},
  volume = {63},
  number = {2},
  pages = {287--313},
  issn = {1077-3142},
  doi = {10.1006/cviu.1996.0020},
  langid = {english}
}

@inproceedings{paradis_color_2020,
  title = {Color {{Normalization}} for {{Robust Automatic Bill}} of {{Materials Generation}} and {{Visual Inspection}} of {{PCBs}}},
  author = {Paradis, Olivia P. and Jessurun, Nathan T. and Tehranipoor, Mark and Asadizanjani, Navid},
  date = {2020-12-01},
  pages = {172--179},
  publisher = {{ASM International}},
  doi = {10.31399/asm.cp.istfa2020p0172},
  url = {https://dl.asminternational.org/istfa/proceedings-abstract/ISTFA2020/83348/172/15412},
  urldate = {2022-12-11},
  eventtitle = {{{ISTFA}} 2020},
  langid = {english}
}

@article{wang_machine_2017,
  title = {A {{Machine Vision Based Automatic Optical Inspection System}} for {{Measuring Drilling Quality}} of {{Printed Circuit Boards}}},
  author = {Wang, Wei-Chien and Chen, Shang-Liang and Chen, L. and Chang, Wan-Jung},
  date = {2017},
  journaltitle = {IEEE Access},
  doi = {10.1109/ACCESS.2016.2631658}
}

@inproceedings{youn_automatic_2014,
  title = {Automatic Classification of {{SMD}} Packages Using Neural Network},
  booktitle = {2014 {{IEEE}}/{{SICE International Symposium}} on {{System Integration}}},
  author = {Youn, SeungGeun and Lee, YounAe and Park, TaeHyung},
  date = {2014-12},
  pages = {790--795},
  doi = {10.1109/SII.2014.7028139},
  eventtitle = {2014 {{IEEE}}/{{SICE International Symposium}} on {{System Integration}}}
}

@article{zhang_convolutional_2018,
  title = {Convolutional Neural Network-Based Multi-Label Classification of {{PCB}} Defects},
  author = {Zhang, Linlin and Jin, Yongqing and Yang, Xuesong and Li, Xia and Duan, Xiaodong and Sun, Yuan and Liu, Hong},
  date = {2018},
  doi = {10.1049/JOE.2018.8279}
}

@inproceedings{zhao_ni_2009,
  title = {{{NI}} Vision Based Automatic Optical Inspection ({{AOI}}) for Surface Mount Devices: {{Devices}} and Method},
  shorttitle = {{{NI}} Vision Based Automatic Optical Inspection ({{AOI}}) for Surface Mount Devices},
  booktitle = {2009 {{International Conference}} on {{Applied Superconductivity}} and {{Electromagnetic Devices}}},
  author = {Zhao, Huibin and Cheng, Jun and Jin, Jianxun},
  date = {2009-09},
  pages = {356--360},
  doi = {10.1109/ASEMD.2009.5306622},
  eventtitle = {2009 {{International Conference}} on {{Applied Superconductivity}} and {{Electromagnetic Devices}}}
}

@article{watkins_inspection_1969,
  title = {Inspection of Integrated Circuit Photomasks with Intensity Spatial Filters},
  author = {Watkins, L.S.},
  date = {1969-09},
  journaltitle = {Proceedings of the IEEE},
  volume = {57},
  number = {9},
  pages = {1634--1639},
  issn = {1558-2256},
  doi = {10.1109/PROC.1969.7348},
  eventtitle = {Proceedings of the {{IEEE}}}
}

@article{tehranipoor_invasion_2017,
	title = {Invasion of the Hardware Snatchers: Cloned Electronics Pollute the Market},
	journal = {IEEE Spectrum},
	author = {Tehranipoor, Mark and Guin, Ujjwal and Bhunia, Swarup},
	month = apr,
	year = {2017},
}

@inproceedings{azhagan_new_2019,
   author = {Mukhil Azhagan and Dhwani Mehta and Hangwei Lu and Sudarshan Agrawal and Praveen Chawla and Mark Tehranipoor and Damon L Woodard and Navid Asadizanjani},
   journal = {International Symposium on Test and Failure Analysis (ISTFA)},
   title = {A New Framework for Automatic Bill of Material Generation and Visual Inspection},
   year = {2019},
}

@misc{newman_anatomy_2020,
	title = {The {Anatomy} of a {Cisco} {Counterfeit} {Shows} {Its} {Dangerous} {Potential}},
	url = {https://www.wired.com/story/counterfeit-cisco-switch-teardown/},
	urldate = {2021-09-12},
	journal = {WIRED},
	author = {Newman, Lily Hay},
	month = jul,
	year = {2020},
}

@misc{noauthor_recognize_nodate,
	title = {Recognize Counterfeit {FB3}-{QS}},
	url = {https://pangolin.com/pages/recognise-counterfeit-fb3-qs},
	author = {{Pangolin Laser Systems}},
}

@ARTICLE{asadi_pcb_2017,
  author={Asadizanjani, Navid and Tehranipoor, Mark and Forte, Domenic},
  journal={IEEE Transactions on Components, Packaging and Manufacturing Technology}, 
  title={PCB Reverse Engineering Using Nondestructive X-ray Tomography and Advanced Image Processing}, 
  year={2017},
  volume={7},
  number={2},
  pages={292-299},
  doi={10.1109/TCPMT.2016.2642824}}

@inproceedings{mcguire_pcb_2019,
	title = {{PCB} {Hardware} {Trojans}: {Attack} {Modes} and {Detection} {Strategies}},
	doi = {10.1109/VTS.2019.8758643},
	booktitle = {2019 {IEEE} 37th {VLSI} {Test} {Symposium} ({VTS})},
	author = {McGuire, Matthew and Ogras, Umit and Ozev, Sule},
	year = {2019},
	pages = {1--6},
}

@misc{robertson_big_2018,
	title = {The Big Hack: How {C}hina Used a Tiny Chip to Infiltrate {U}.{S}. Companies},
	url = {https://www.bloomberg.com/news/features/2018-10-04/the-big-hack-how-china-used-a-tiny-chip-to-infiltrate-america-s-top-companies},
	urldate = {2020-11-12},
	author = {Robertson, J. and Riley, M.},
	year = {2018},
	publication = {Bloomberg}
}

@article{grow_dangerous_2008,
author={Grow,Brian and Chi-Chu Tschang and Edwards,Cliff and Burnsed,Brian and Epstein,Keith},
year={2008},
month=Oct,
day=13,
title={Dangerous fakes},
journal={Bloomberg Businessweek},
number={4103}
}

@article{guin_a_comprehensive_2014,
	author = {Guin, Ujjwal and DiMase, Daniel and Tehranipoor, Mohammad},
	id = {Guin2014},
	journal = {Journal of Electronic Testing},
	number = {1},
	pages = {25--40},
	title = {A Comprehensive Framework for Counterfeit Defect Coverage Analysis and Detection Assessment},
	url = {https://doi.org/10.1007/s10836-013-5428-2},
	volume = {30},
	year = {2014},
	month = Feb
}

@inproceedings{goetz_counterfeit_2017,
	title = {Counterfeit {Electronic} {Components} {Identification}: {A} {Case} {Study}},
	author = {Goetz, Martin and Varma, Ramesh},
	year = {2017},
}

@misc{senate_investigation_2012,
	title = {Investigation into counterfeit electronic parts in the department of defense supply chain},
	month = nov,
	year = {2012},
}

@article{guin_counterfeit_2014,
author = {Guin, Ujjwal and Huang, Ke and DiMase, Daniel and Carulli, John M and Tehranipoor, Mohammad and Makris, Yiorgos},
doi = {10.1109/JPROC.2014.2332291},
journal = {Proc. IEEE},
month = aug,
number = {8},
pages = {1207--1228},
publisher = {IEEE},
title = {Counterfeit Integrated Circuits: A Rising Threat in the Global Semiconductor Supply Chain},
volume = {102},
year = {2014}
}

@article{appelbaum_nsas_2013,
	title = {{NSA}'s {Secret} {Toolbox}: {Unit} {Offers} {Spy} {Gadgets} for {Every} {Need}},
	url = {https://www.spiegel.de/international/world/nsa-secret-toolbox-ant-unit-offers-spy-gadgets-for-every-need-a-941006.html},
	journal = {Der Spiegel},
	author = {Appelbaum, Jacob and HORCHERT, JUDITH and REISSMANN, OLE and ROSENBACH, MARCEL and SCHINDLER, JÖRG and STÖCKER, CHRISTIAN},
	month = dec,
	year = {2013},
}

@article{harrison_malicious_2021,
	title = {On malicious implants in {PCBs} throughout the supply chain},
	volume = {79},
	issn = {0167-9260},
	url = {https://www.sciencedirect.com/science/article/pii/S0167926021000304},
	doi = {https://doi.org/10.1016/j.vlsi.2021.03.002},
	journal = {Integration},
	author = {Harrison, Jacob and Asadizanjani, Navid and Tehranipoor, Mark},
	year = {2021},
	pages = {12--22},
}

@inproceedings{chen_addressing_2017,
  title = {Addressing Supply Chain Risks of Microelectronic Devices through Computer Vision},
  booktitle = {2017 {{IEEE Applied Imagery Pattern Recognition Workshop}} ({{AIPR}})},
  author = {Chen, Zhenhua and Wanyan, Tingyi and Rao, Ramya and Cutilli, Benjamin and Sowinski, James and Crandall, David and Templeman, Robert},
  date = {2017-10},
  pages = {1--8},
  publisher = {{IEEE}},
  location = {{Washington, DC, USA}},
  doi = {10.1109/AIPR.2017.8457956},
  eventtitle = {2017 {{IEEE Applied Imagery Pattern Recognition Workshop}} ({{AIPR}})},
  isbn = {978-1-5386-1235-4},
  langid = {english}
}

@article{pramerdorfer_dataset_2015,
  title = {A Dataset for Computer-Vision-Based {{PCB}} Analysis},
  author = {Pramerdorfer, C. and Kampel, M.},
  date = {2015},
  journaltitle = {2015 14th IAPR International Conference on Machine Vision Applications (MVA)},
  doi = {10.1109/MVA.2015.7153209}
}

@online{huang_pcb_2019,
  title = {A {{PCB}} Dataset for Defects Detection and Classification},
  author = {Huang, W. and Wei, P.},
  date = {2019},
  eprint = {1901.08204},
  eprinttype = {arxiv},
  publisher = {{arxiv.org}},
  url = {https://arxiv.org/abs/1901.08204},
  archiveprefix = {arXiv}
}

@article{mahalingam_pcbmetal_2019,
  title = {{{PCB}}-{{METAL}}: A {{PCB Image Dataset}} for {{Advanced Computer Vision Machine Learning Component Analysis}}},
  shorttitle = {{{PCB}}-{{METAL}}},
  author = {Mahalingam, Gayathri and Gay, Kevin and Ricanek, K.},
  date = {2019},
  journaltitle = {2019 16th International Conference on Machine Vision Applications (MVA)},
  doi = {10.23919/MVA.2019.8757928}
}

@inproceedings{gang_coresets_2020,
  title = {Coresets for {{PCB Character Recognition}} Based on {{Deep Learning}}},
  booktitle = {2020 {{International Conference}} on {{Artificial Intelligence}} in {{Information}} and {{Communication}} ({{ICAIIC}})},
  author = {Gang, Sumyung and Fabrice, Ndayishimiye and Lee, JoonJae},
  date = {2020-02},
  pages = {637--642},
  doi = {10.1109/ICAIIC48513.2020.9065271},
  eventtitle = {2020 {{International Conference}} on {{Artificial Intelligence}} in {{Information}} and {{Communication}} ({{ICAIIC}})}
}

@online{tang_online_2019,
  title = {Online {{PCB Defect Detector On A New PCB Defect Dataset}}},
  author = {Tang, Sanli and He, Fan and Huang, Xiaolin and Yang, Jie},
  date = {2019-02-16},
  eprint = {1902.06197},
  eprinttype = {arxiv},
  primaryclass = {cs},
  url = {http://arxiv.org/abs/1902.06197},
  urldate = {2021-11-18},
  archiveprefix = {arXiv}
}

@article{gang_character_2021,
  title = {Character {{Recognition}} of {{Components Mounted}} on {{Printed Circuit Board Using Deep Learning}}},
  author = {Gang, S. and Fabrice, Ndayishimiye and Chung, Daewon and Lee, JoonJae},
  date = {2021},
  journaltitle = {Sensors},
  doi = {10.3390/s21092921}
}

@article{reza_icchipnet_2020,
  title = {{{IC}}-{{ChipNet}}: Deep {{Embedding Learning}} for {{Fine}}-Grained {{Retrieval}}, {{Recognition}}, and {{Verification}} of {{Microelectronic Images}}},
  shorttitle = {{{IC}}-{{ChipNet}}},
  author = {Reza, Md Alimoor and Crandall, David J.},
  date = {2020},
  journaltitle = {2020 IEEE Applied Imagery Pattern Recognition Workshop (AIPR)},
  doi = {10.1109/AIPR50011.2020.9425131}
}

@report{lu_ficspcb_2020,
  title = {{{FICS}}-{{PCB}}: A {{Multi}}-{{Modal Image Dataset}} for {{Automated Printed Circuit Board Visual Inspection}}},
  shorttitle = {{{FICS}}-{{PCB}}},
  author = {Lu, Hangwei and Mehta, Dhwani and Paradis, Olivia and Asadizanjani, Navid and Tehranipoor, Mark and Woodard, Damon L.},
  date = {2020},
  number = {366},
  url = {http://eprint.iacr.org/2020/366},
  urldate = {2021-11-18}
}

@article{fridman_changechip_2021,
  title={ChangeChip: A Reference-Based Unsupervised Change Detection for PCB Defect Detection},
  url={http://arxiv.org/abs/2109.05746},
  note={arXiv: 2109.05746},
  journal={arXiv:2109.05746 [cs]},
  author={Fridman, Yehonatan and Rusanovsky, Matan and Oren, Gal},
  year={2021},
  month=Sep
}

@misc{karanth_pcbexperiment_2020,
  title={PCBexperiment},
  url={https://kaggle.com/namrathakaranth/pcbexperiment},
  journal={Kaggle},
  author={Karanth, Namratha},
  year={2020},
  month=Aug
}

@article{li_multisensor_2021,
  title = {Multisensor {{Image Fusion}} for {{Automated Detection}} of {{Defects}} in {{Printed Circuit Boards}}},
  author = {Li, Mengke and Yao, Naifu and Liu, Sha and Li, Shouqing and Zhao, Yongqiang and Kong, Seong G.},
  date = {2021-10},
  journaltitle = {IEEE Sensors Journal},
  volume = {21},
  number = {20},
  pages = {23390--23399},
  issn = {1558-1748},
  doi = {10.1109/JSEN.2021.3106057},
  eventtitle = {{{IEEE Sensors Journal}}},
}

@article{shieh_applying_2021,
  title = {Applying Deep Learning to Defect Detection in Printed Circuit Boards via a Newest Model of You-Only-Look-Once},
  author = {Shieh, J.-S.},
  date = {2021-05-21},
  journaltitle = {Mathematical Biosciences and Engineering},
  volume = {18},
  number = {4},
  pages = {4411--4428},
  doi = {10.3934/mbe.2021223}
}

@online{ganapathy_defect_2021,
  title = {Defect Detection and Classification in Manufacturing Using {{Amazon Lookout}} for {{Vision}} and {{Amazon Rekognition Custom Labels}}},
  author = {Ganapathy, Prashanth and Gupta, Amit},
  date = {2021-07-13T08:01:33-08:00},
  url = {https://aws.amazon.com/blogs/machine-learning/defect-detection-and-classification-in-manufacturing-using-amazon-lookout-for-vision-and-amazon-rekognition-custom-labels/},
  urldate = {2021-12-02},
  langid = {american},
  organization = {{Amazon Web Services}}
}

@article{wei_cnnbased_2018,
  title = {{{CNN-based}} Reference Comparison Method for Classifying Bare {{PCB}} Defects},
  author = {Wei, Peng and Liu, Chang and Liu, M. and Gao, Yunlong and Liu, Hong},
  date = {2018},
  doi = {10.1049/JOE.2018.8271}
}

@inproceedings{wu_automated_2010,
  title = {Automated Visual Inspection of Surface Mounted Chip Components},
  booktitle = {2010 {{IEEE International Conference}} on {{Mechatronics}} and {{Automation}}},
  author = {Wu, Huihui. and Feng, Guanglin. and Li, Huiwen. and Zeng, Xianrong.},
  date = {2010-08},
  pages = {1789--1794},
  issn = {2152-744X},
  doi = {10.1109/ICMA.2010.5588029},
  eventtitle = {2010 {{IEEE International Conference}} on {{Mechatronics}} and {{Automation}}}
}

@inproceedings{richter_development_2017,
  title = {On the Development of Intelligent Optical Inspections},
  booktitle = {2017 {{IEEE}} 7th {{Annual Computing}} and {{Communication Workshop}} and {{Conference}} ({{CCWC}})},
  author = {Richter, Johannes and Streitferdt, Detlef and Rozova, Elena},
  date = {2017-01},
  pages = {1--6},
  doi = {10.1109/CCWC.2017.7868455},
  eventtitle = {2017 {{IEEE}} 7th {{Annual Computing}} and {{Communication Workshop}} and {{Conference}} ({{CCWC}})},
}

@inproceedings{mazondeoliveira_detecting_2017,
  title = {Detecting {{Modifications}} in {{Printed Circuit Boards}} from {{Fuel Pump Controllers}}},
  booktitle = {2017 30th {{SIBGRAPI Conference}} on {{Graphics}}, {{Patterns}} and {{Images}} ({{SIBGRAPI}})},
  author = {Mazon De Oliveira, Thomas Jose and Wehrmeister, Marco Aurelio and Nassu, Bogdan Tomoyuki},
  date = {2017-10},
  pages = {87--94},
  issn = {2377-5416},
  doi = {10.1109/SIBGRAPI.2017.18},
  eventtitle = {2017 30th {{SIBGRAPI Conference}} on {{Graphics}}, {{Patterns}} and {{Images}} ({{SIBGRAPI}})}
}

@article{wang_machine_2013,
  title = {Machine {{Vision-Based Defect Detection}} in {{IC Images Using}} the {{Partial Information Correlation Coefficient}}},
  author = {Wang, Chien-Chih and Jiang, Bernard C. and Lin, Jing-You and Chu, Chien-Cheng},
  date = {2013-08},
  journaltitle = {IEEE Transactions on Semiconductor Manufacturing},
  volume = {26},
  number = {3},
  pages = {378--384},
  issn = {1558-2345},
  doi = {10.1109/TSM.2013.2261566},
  eventtitle = {{{IEEE Transactions}} on {{Semiconductor Manufacturing}}}
}

@book{lin_using_2019,
  type = {10.1007/s10489-019-01486-5},
  title = {Using Convolutional Neural Networks for Character Verification on Integrated Circuit Components of Printed Circuit Boards},
  author = {Lin, C. H. and Wang, S. H. and Lin, C. J.},
  date = {2019},
  journaltitle = {Applied Intelligence},
  publisher = {{Springer}},
  url = {https://link.springer.com/article/10.1007/s10489-019-01486-5}
}

@article{larochelle_exploring_2009,
  title = {Exploring {{Strategies}} for {{Training Deep Neural Networks}}},
  author = {Larochelle, Hugo and Bengio, Yoshua and Louradour, Jerome and Lamblin, Pascal},
  date = {2009},
  journaltitle = {Journal of machine learning research},
  volume = {10},
  number = {1},
  pages = {40},
  langid = {english},
}

@online{garcia-garcia_review_2017,
  title = {A {{Review}} on {{Deep Learning Techniques Applied}} to {{Semantic Segmentation}}},
  author = {Garcia-Garcia, Alberto and Orts-Escolano, Sergio and Oprea, Sergiu and Villena-Martinez, Victor and Garcia-Rodriguez, Jose},
  date = {2017-04-22},
  eprint = {1704.06857},
  eprinttype = {arxiv},
  primaryclass = {cs},
  url = {http://arxiv.org/abs/1704.06857},
  urldate = {2022-01-08},
  archiveprefix = {arXiv},
}

@inproceedings{long_fully_2015,
  title = {Fully {{Convolutional Networks}} for {{Semantic Segmentation}}},
  author = {Long, Jonathan and Shelhamer, Evan and Darrell, Trevor},
  date = {2015},
  pages = {3431--3440},
  url = {https://openaccess.thecvf.com/content_cvpr_2015/html/Long_Fully_Convolutional_Networks_2015_CVPR_paper.html},
  urldate = {2022-01-08},
  eventtitle = {Proceedings of the {{IEEE Conference}} on {{Computer Vision}} and {{Pattern Recognition}}},
}

@inproceedings{wang_understanding_2018,
  title = {Understanding {{Convolution}} for {{Semantic Segmentation}}},
  booktitle = {2018 {{IEEE Winter Conference}} on {{Applications}} of {{Computer Vision}} ({{WACV}})},
  author = {Wang, Panqu and Chen, Pengfei and Yuan, Ye and Liu, Ding and Huang, Zehua and Hou, Xiaodi and Cottrell, Garrison},
  date = {2018-03},
  pages = {1451--1460},
  doi = {10.1109/WACV.2018.00163},
  eventtitle = {2018 {{IEEE Winter Conference}} on {{Applications}} of {{Computer Vision}} ({{WACV}})},
}

@inproceedings{yu_bisenet_2018,
  title = {{{BiSeNet}}: {{Bilateral Segmentation Network}} for {{Real-time Semantic Segmentation}}},
  shorttitle = {{{BiSeNet}}},
  author = {Yu, Changqian and Wang, Jingbo and Peng, Chao and Gao, Changxin and Yu, Gang and Sang, Nong},
  date = {2018},
  pages = {325--341},
  url = {https://openaccess.thecvf.com/content_ECCV_2018/html/Changqian_Yu_BiSeNet_Bilateral_Segmentation_ECCV_2018_paper.html},
  urldate = {2022-01-08},
  eventtitle = {Proceedings of the {{European Conference}} on {{Computer Vision}} ({{ECCV}})},
}

@inproceedings{zhang_context_2018,
  title = {Context {{Encoding}} for {{Semantic Segmentation}}},
  booktitle = {2018 {{IEEE}}/{{CVF Conference}} on {{Computer Vision}} and {{Pattern Recognition}}},
  author = {Zhang, Hang and Dana, Kristin and Shi, Jianping and Zhang, Zhongyue and Wang, Xiaogang and Tyagi, Ambrish and Agrawal, Amit},
  date = {2018-06},
  pages = {7151--7160},
  publisher = {{IEEE}},
  location = {{Salt Lake City, UT, USA}},
  doi = {10.1109/CVPR.2018.00747},
  url = {https://ieeexplore.ieee.org/document/8578845/},
  urldate = {2022-01-08},
  eventtitle = {2018 {{IEEE}}/{{CVF Conference}} on {{Computer Vision}} and {{Pattern Recognition}} ({{CVPR}})},
  isbn = {978-1-5386-6420-9},
  langid = {english},
}

@inproceedings{paul_gomac,
	author = {Paul Calzada and Jacob Harrison and Navid Asadizanjani and Mark Tehranipoor and Praveen Chawla},
	title = {{PCB} Trojan Detection using Optical Imaging},
	booktitle = {46th GOMAC Tech},
	year = {2022},
	month = Mar,
	address = {Miami, FL}
}

@misc{mehta_fics_2022,
  title = {{{FICS PCB X-ray}}: {{A}} Dataset for Automated Printed Circuit Board Inter-Layers Inspection},
  shorttitle = {{{FICS PCB X-ray}}},
  author = {Mehta, Dhwani and True, John and Dizon-Paradis, Olivia P. and Jessurun, Nathan and Woodard, Damon L. and Asadizanjani, Navid and Tehranipoor, Mark},
  date = {2022},
  number = {924},
  url = {https://eprint.iacr.org/2022/924},
  urldate = {2022-12-11}
}
\printacronyms
\end{document}